\documentclass[runningheads]{llncs}
\usepackage{graphicx}
\usepackage{amsmath,amssymb} % define this before the line numbering.
\usepackage{color}
\usepackage{epsfig}
\usepackage{algorithm}
\usepackage{fancybox}
\usepackage[noend]{algpseudocode}
\usepackage{caption}  %JWL: gives some warning in latex compilation.
\usepackage{subcaption}
\usepackage[width=122mm,left=12mm,paperwidth=146mm,height=193mm,top=12mm,paperheight=217mm]{geometry}
\begin{document}
% \renewcommand\thelinenumber{\color[rgb]{0.2,0.5,0.8}\normalfont\sffamily\scriptsize\arabic{linenumber}\color[rgb]{0,0,0}}
% \renewcommand\makeLineNumber {\hss\thelinenumber\ \hspace{6mm} \rlap{\hskip\textwidth\ \hspace{6.5mm}\thelinenumber}}
% \linenumbers
\pagestyle{headings}
\mainmatter
\def\ECCV16SubNumber{975}  % Insert your submission number here

\title{Ego2Top: Matching Viewers in Egocentric and Top-view Videos} % Replace with your title

\titlerunning{Ego2Top: Matching Viewers in Egocentric and Top-view Videos}

\authorrunning{Shervin Ardeshir and Ali Borji}

\author{Shervin Ardeshir and Ali Borji}
\institute{Center for Research in Computer Vision\\
	University of Central Florida\\
	\email{ ardeshir@cs.ucf.edu,
		aborji@crcv.ucf.edu}
}

\maketitle
\begin{abstract}
Egocentric cameras are becoming increasingly popular and provide us with large amounts of videos, captured from the first person perspective. At the same time, surveillance cameras and drones offer an abundance of visual information, often captured from top-view. Although these two sources of information have been separately studied in the past, they have not been collectively studied and related. Having a set of egocentric cameras and a top-view camera capturing the same area, we propose a framework to identify the egocentric viewers in the top-view video. We utilize two types of features for our assignment procedure. Unary features encode what a viewer (seen from top-view or recording an egocentric video) visually experiences over time. Pairwise features encode the relationship between the visual content of a pair of viewers. Modeling each view (egocentric or top) by a graph, the assignment process is formulated as spectral graph matching. Evaluating our method over a dataset of 50 top-view and 188 egocentric videos taken in different scenarios demonstrates the efficiency of the proposed approach in assigning egocentric viewers to identities present in top-view camera. We also study the effect of different parameters such as the number of egocentric viewers and visual features. An extended abstract of this effort has been reported in \cite{ardeshiregocentric}.
\end{abstract}
\section{Introduction}
The availability of large amounts of egocentric videos captured by cellphones and wearable devices such as GoPro cameras and Google Glass has opened the door to a lot of interesting research in computer vision \cite{egoActionFathi,egoDailyAction,egoFOVLocalization}. At the same time, videos captured with top-down static cameras such as surveillance cameras in airports and subways, unmanned aerial vehicles (UAVs) and drones, provide us with a lot of invaluable information about activities and events taking place at different locations and environments. Relating these two complementary, but drastically different sources of visual information can provide us with rich analytical power, and help us explore what can not be inferred from each of these sources taken separately. Establishing such a relationship can have several applications. For example, athletes can be equipped with body-worn cameras, and their egocentric videos together with the top-view videos can offer new data useful for better technical and tactical sport analysis. Moreover, due to the use of wearable devices and cameras by law enforcement officers, finding the person behind an egocentric camera in a surveillance network could be a useful application. Furthermore, fusing these two types of information can result in better 3D reconstruction of an environment by combining the top-view information with first person views. 

\begin{figure}[t]
	\centering
	\includegraphics[width=1\linewidth]{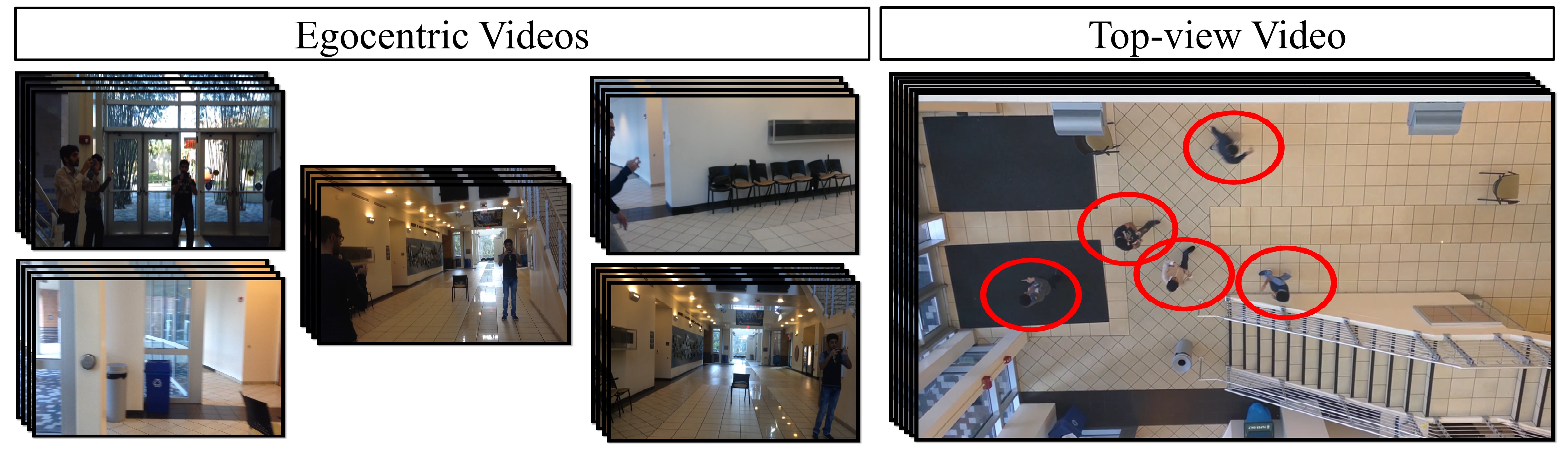}
	\caption{Left shows a set of 5 egocentric videos. Right shows a top-view video capturing the whole scene. The viewers are highlighted using red circles in the top-view video. We aim to answer the two following questions: 1) Does this set of egocentric videos belong to the viewers visible in the top-view video? 2) Assuming they do, which viewer is capturing which egocentric video?}	
	\label{fig:Example}
\end{figure}

The first necessary step to utilize information from these two sources, is to establish correspondences between the two views. In other words, a matching between egocentric cameras and the people present in the top-view camera is needed. In this effort, we attempt to address this problem. More specifically, our goal is to localize people recording egocentric videos, in a top-view reference camera. To the best of our knowledge, such an effort has not been done so far. In order to evaluate our method, we designed the following setup. A dataset containing several test cases is collected. In each test case, multiple people were asked to move freely in a certain environment and record egocentric videos. We refer to these people as ego-centric \textit{viewers}. At the same time, a top-view camera was recording the entire scene/area including all the egocentric viewers and possibly other intruders. An example case is illustrated in Figure \ref{fig:Example}. 

Given a set of egocentric videos and a top-view surveillance video, we try to answer the following two questions: 1) Does this set of egocentric videos belong to the viewers visible in the top-view camera? 2) If yes, then which viewer is capturing which egocentric video? 
To answer these questions, we need to compare a set of egocentric videos to a set of viewers visible in a single top-view video. 
To find a matching, each set is represented by a graph and the two graphs are compared using a spectral graph matching technique \cite{spectralMatching}. In general, this problem can be very challenging due to the nature of egocentric cameras. Since the camera-holder is not visible in his own egocentric video leaving us with no cues about his visual appearance. 

In what follows we briefly mention some challenges concerning this problem and sketch the layout of our approach.% Note that an egocentric video captures a person's field of view rather than his spatial location. Therefore, the content of a viewer's egocentric video corresponds to the content of the viewer's expected field of view in the top-view camera. These contents might not be necessarily of the same nature. For example, the egocentric viewer observes other people or objects from the side view while the top-view camera captures the scene from a different perspective. Further, there might be contents that only one of the cameras captures (due to occlusion, shadow, etc). Due to that nature, having a direct comparison between an egocentric video and a viewer is not meaningful. Hence, we attempt to discover features describing a set of egocentric videos which is preserved across the view-points.

\begin{figure}[t]
	\centering
	\includegraphics[width=1\linewidth]{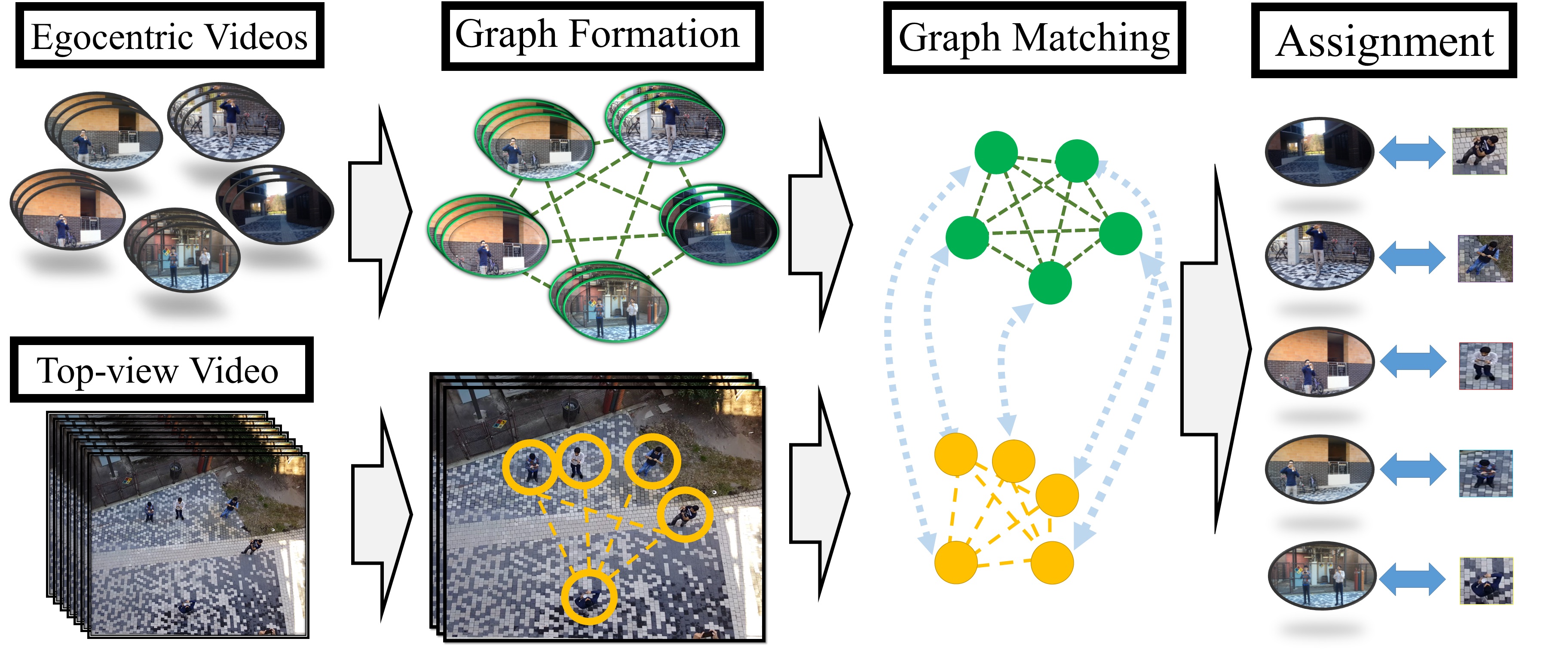}
	\caption{The input to our framework is a set of egocentric videos (in this case 5 videos), and one top-view video. The goal is defined as assigning the egocentric videos to the people recording them. A graph is formed on the set of egocentric videos (each node being one of the egocentric videos), and the other graph is formed on the top-view video (each node being one of the targets present in the video). Using spectral graph matching, a soft assignment is found between the two graphs, and using a soft-to-hard assignment, each egocentric video is assigned to one of the viewers in the top-view video. This assignment is our answer to the second question in \ref{fig:Example}.}
	\label{fig:teaser}
\end{figure}

In order to have an understanding of the behavior of each individual in the top-view video, we use a multiple object tracking method \cite{theWayTheyMove} to extract the viewer's trajectory in the top-view video. Note that an egocentric video captures a person's field of view rather than his spatial location. Therefore, the content of a viewer's egocentric video, a 2D scene, corresponds to the content of the viewer's field of view in the top-view camera. For the sake of brevity, we refer to a viwer's top-view field of view as Top-FOV in what follows. Since trajectories computed by multiple object tracking do not provide us with the orientation of the egocentric cameras in the top-view video, we employ the assumption that for the most part humans tend to look straight ahead and therefore shoot videos from the visual content in front of them. Note that this is not a restrictive assumption as most ego-centric cameras are body worn (Please see Figure \ref{fig:inFOVs}). Having an estimate of a viewer's orientation and Top-FOV, we then encode the changes in his Top-FOV over time and use it as a descriptor. We show that this feature correlates with the change in the global visual content (or Gist) of the scene observed in his corresponding egocentric video. 

We also define pairwise features to capture the relationship between two egocentric videos, and also the relationship between two viewers in the top-view camera. Intuitively, if an egocentric viewer observes a certain scene and another egocentric viewer comes across the same scene later, this could hint as a relationship between the two cameras. If we match a top-view viewer to one of the two egocentric videos, we are likely to be able to find the other viewer using the mentioned relationship. As we experimentally show, this pairwise relationship significantly improves our assignment accuracy. This assignment will lead to defining a score measuring the similarity between the two graphs. Our experiments demonstrate that the graph matching score could be used for verifying if the top-view video is in fact, capturing the egocentric viewers (See the diagram shown in Figure \ref{fig:sceneMatching_teaser}). 

The rest of this work is as follows. In section \ref{sec:related_work}, we mention related works to our study. In section \ref{sec:framework}, we describe the details of our framework. Section \ref{sec:results} presents our experimental results followed by discussions and conclusions in Section \ref{sec:conclusion}.
%\begin{figure}
%	\begin{center}
%		\begin{subfigure}{0.32\textwidth}
%			\includegraphics[width=\textwidth]{Figures/Examples/00000.png}	
%		\end{subfigure}
%		\begin{subfigure}{0.32\textwidth}
%			\includegraphics[width=\textwidth]{Figures/Examples/00001.png}
%		\end{subfigure}
%		\begin{subfigure}{0.32\textwidth}
%			\includegraphics[width=\textwidth]{Figures/Examples/00002.png}
%		\end{subfigure}		
%		\caption{}	
%		\label{fig:Dataset}			
%	\end{center}
%\end{figure}
\begin{figure}[t]
	\begin{center}
		\includegraphics[width=.7\textwidth]{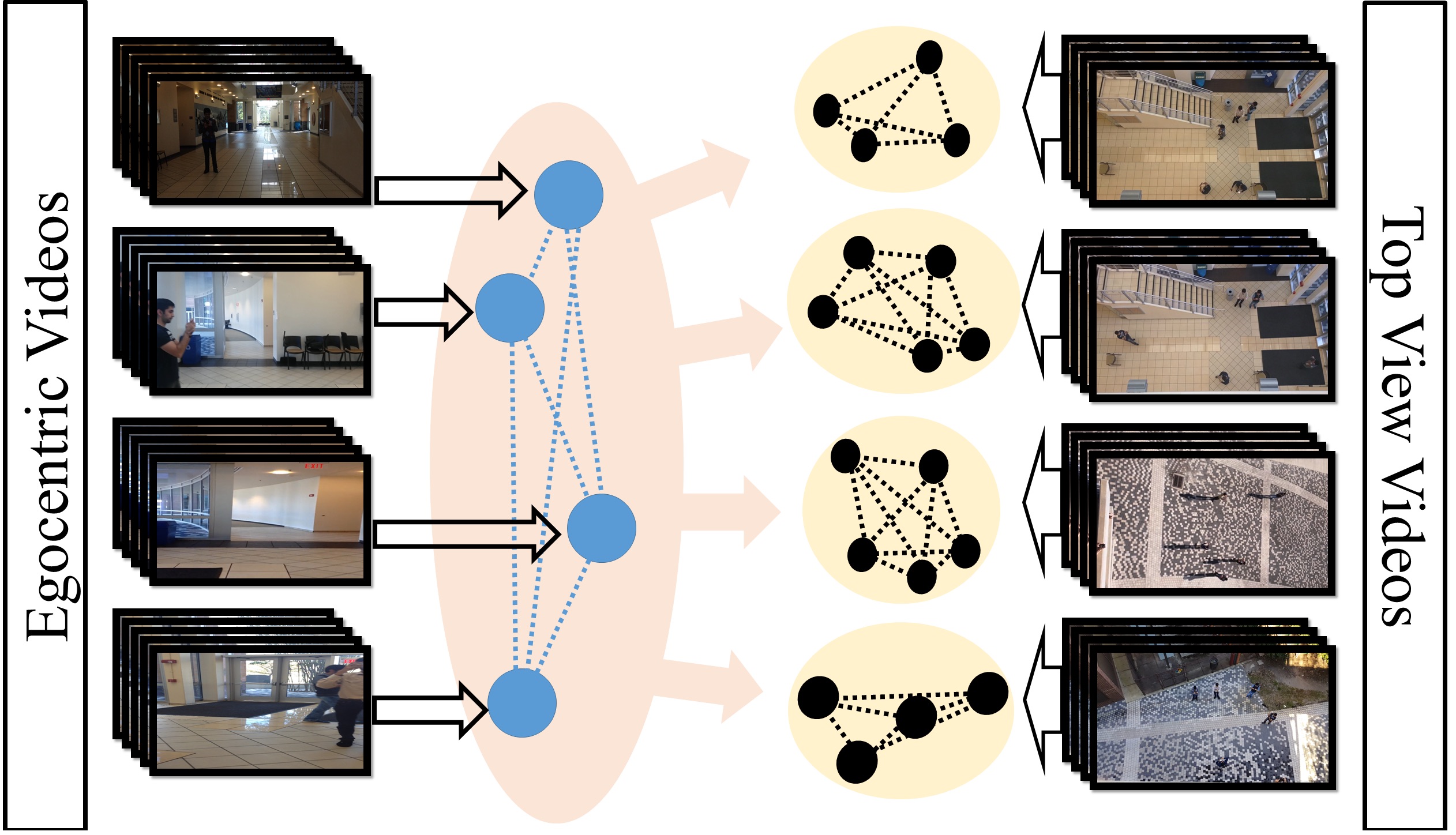}			
		\caption{Adapting our method for evaluating top-view videos. We compare the graph formed on the set of egocentric videos to the ones built on each top-view video. The top-view videos are then ranked based on the graph matching similarity. The performance of this ranking gives us insight on our first question.}
		\label{fig:sceneMatching}			
	\end{center}
\end{figure}
\section{Related Work}
\label{sec:related_work}
Visual analysis of egocentric videos has recently became a hot topic in computer vision~\cite{egoKanade,egoEvolutionSurvey}, from recognizing daily activities \cite{egoDailyAction,egoActionFathi} to object detection \cite{egoObjectDetection}, video summarization \cite{egoVideoSummarization}, and predicting gaze behavior~\cite{egoli2013learning,egoPolatsekNovelty,Borji2014look}. In the following, we review some previous work related to ours spanning \textit{Relating static and egocentric},~\textit{Social interactions among egocentric viewers}, and~\textit{Person identification and localization}.\\

\noindent\textbf{Relating Static and Egocentric Cameras:} Some studies have addressed relationships between moving and static cameras. 
Interesting works reported in \cite{egoMobileFixedObjectDetection,egoMobileFixedMasterSlave} have studied the relationship between mobile and static cameras for the purpose of improving object detection accuracy. \cite{egoExo} fuses information from egocentric and exocentric vision (other cameras in the environment) and laser depth range data to improve depth perception in 3D reconstruction. \cite{egoPredictingGaze} predicts gaze behavior in social scenes using first-person and third-person cameras. \\

\noindent\textbf{Social Interactions among Egocentric Viewers:}
To explore the relationship among multiple egocentric viewers, \cite{egoWisdomOfTheCrowd} combines several egocentric videos to achieve a more complete video with less quality degradation by estimating the importance of different scene regions and incorporating the consensus among several egocentric videos. Fathi et al.,~\cite{egoSocialInteractions} detect and recognize the type of social interactions such as dialogue, monologue, and discussion by detecting human faces and estimating their body and head orientations. \cite{egoMultiTaskClustering} proposes a multi-task clustering framework, which searches for coherent clusters of daily actions using the notion that people tend to perform similar actions in certain environments such as workplace or kitchen. \cite{egoYouDoILearn} proposes a framework that discovers static and movable objects used by a set of egocentric users.\\

\noindent\textbf{Person Identification and Localization:} Perhaps, the most similar computer vision task to ours is person re-identification \cite{reidCPS,reidReimannian,reidSDALF}. The objective here is to find the person present in one static camera, in another overlapping or non-overlapping static camera. However, the main cue in human re-identification is visual appearance of humans, which is absent in egocentric videos. 
Tasks such as human-identification and localization in egocentric cameras have been studied in the past. \cite{egoHeadMotion} uses the head motion of an egocentric viewer as a biometric signature for determine which videos have been captured by the same person. \cite{egoSurfing} identifies egocentric observers in other egocentric videos, using their head motion. 
Relating geo-spatial location to user shared visual content has also been explored. \cite{egoFOVLocalization} localizes the field of view of an egocentric camera by matching it against a reference dataset of videos or images (such as Google street view), and \cite{TagrefinementCVPR14} refines the geo-location of images by matching them against user shared images. Landmarks and map symbols are used in~\cite{egoWhereAmI} to perform self localization on the map. \cite{GISobject} use semantic cues for spatial localization, and \cite{Geosemantic} uses location information to infer semantic information. 
%[[talk about im2pos, zahir, james hayes,ccc]]
\section{Framework}
\label{sec:framework}
The block diagram in Figure \ref{fig:teaser} illustrates different steps of our approach. 
First, each view (ego-centric or top-down) is represented by a graph which defines the relationship among the viewers present in the scene. These two graphs may not have the same number of nodes as some the egocentric videos might not be available, or some individuals present in the top-view video might not be capturing videos. Each graph consists of a set of nodes, each of which represents one viewer (egocentric or top-view), and the edges of the graph encode pairwise relationships between pairs of viewers.

We represent each viewer in top-view by describing his expected Top-FOV, and in egocentric view by the visual content of his video over time. This description is encoded in the nodes of the graphs. We also define pairwise relationships between pairs of viewers, which is encoded as the edge features of the graph (i.e., how two viewers' visual experience relate to each other).

Second, we use spectral graph matching to compute a score measuring the similarity between the two graphs, alongside with an assignment from the nodes of the egocentric graph to the nodes of the top-view graph. 

Our experiments show that the graph matching score can be used as a measure of similarity between the egocentric graph and the top-view graph. Therefore, it can be used as a measure for verifying if a set of egocentric videos have been shot in the same environment captured by the top-view camera. In other words, we can evaluate the capability of our method in terms of answering our first question. In addition, the assignment obtained by the graph matching suggests an answer to our second question. We organize this section by going over the graph formation process for each of the views, and then describing the details of the matching procedure.

%\begin{figure}
%	\centering
%	\includegraphics[width=1\linewidth]{Figures/block_diagram_v3.png}
%	\caption{Block Diagram.}
%	\label{fig:block_diag}
%\end{figure}

\subsection{Graph Representation} Each view (egocentric or top-view) is described using a single graph. 
The set of egocentric videos is represented using a graph in which each node represents one of the egocentric videos, and an edge captures the pairwise relationship between the content of the two videos. 

In the top-view graph, each node represents the visual experience of a viewer being tracked (in the top-view camera), and an edge captures the pairwise relationship between the two. By visual experience we mean what a viewer is expected to observe during the course of his recording seen from the top view.\\
    
\noindent \textbf{3.1.1 \ \ Modeling the Top-View Graph:} In order to model the visual experience of a viewer in a top-view camera, we need to have knowledge about his spatial location (trajectory) throughout the video. We employ the multiple object tracking method presented in \cite{theWayTheyMove} and extract a set of trajectories, each corresponding to one of the viewers in the scene. Similar to \cite{theWayTheyMove}, we use annotated bounding boxes, and provide their centers as an input to the multiple object tracker. Our tracking results are nearly perfect due to several reasons: the high quality of videos, high video frame rate, and lack of challenges such as occlusion in the top-view videos.  
 
\begin{figure}[t]
\begin{center}
			\begin{subfigure}{0.33\textwidth}
				\includegraphics[width=\textwidth]{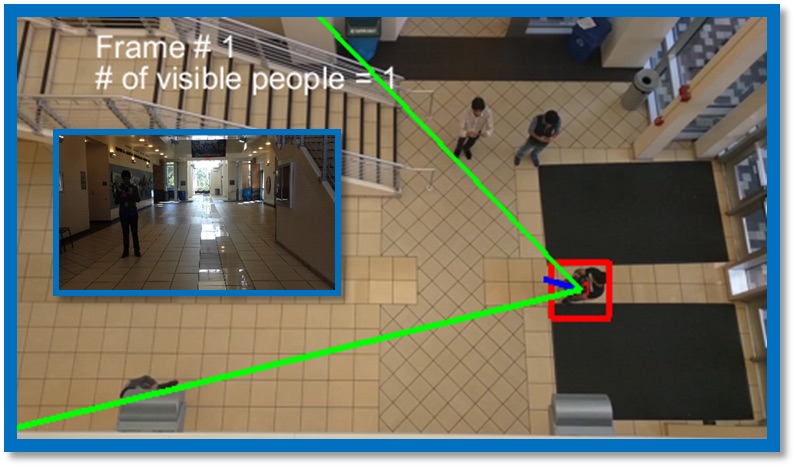}				\caption{}	
			\end{subfigure}\hfill	
			\begin{subfigure}{0.33\textwidth}
				\includegraphics[width=\textwidth]{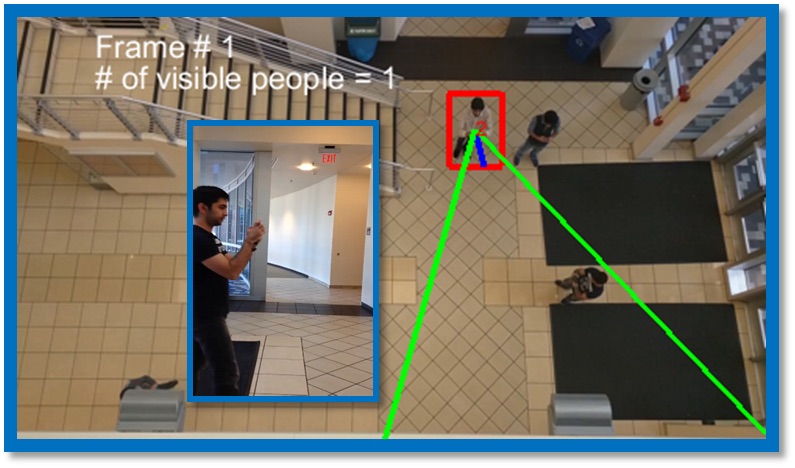}	
				\caption{}		
			\end{subfigure}\hfill	
			\begin{subfigure}{0.33\textwidth}
				\includegraphics[width=\textwidth]{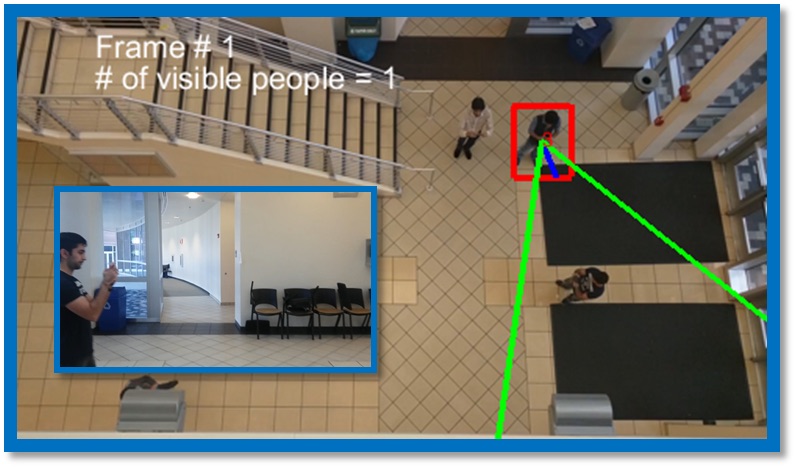}	
				\caption{}	
			\end{subfigure}\hfill									
	\caption{Expected field of view for three different viewers in the top-view video alongside with their corresponding egocentric frames. The short dark blue line shows the estimated orientation of the camera. The Top-FOV shown in (b) and (c) have a high overlap, therefore we expect their egocentric videos to have relatively similar visual content compared to the pairs (a,b) or (a,c) at this specific time.}
	\label{fig:inFOVs}			
\end{center}	
\end{figure}

Each node represents one of the individuals being tracked. Employing the general assumption that people often tend to look straight ahead, we use a person's speed vector as the direction of his camera at time t (denoted as $\theta_t$). Further, assuming a fixed angle ($\theta_d$), we expect the content of the person's egocentric video to be consistent with the content included in a 2D cone formed by the two rays emanating from the viewer's location and with angles $\theta - \theta_d$ and $\theta + \theta_d$. Figure \ref{fig:inFOVs} illustrates the expected Top-FOV for three different individuals present in a frame. In our experiments, we set $\theta_d$ to 30 degrees. In theory, angle $\theta_d$ can be estimated more accurately by knowing intrinsic camera parameters such as focal length and sensor size of the corresponding egocentric camera. However, since we do not know the corresponding egocentric camera, we set it to a default value.

Top-FOVs are not directly comparable to viewers' egocentric views. The area in the Top-FOV in a top-view video mostly contains the ground floor which is not what an ego-centric viewer usually observes in front of him. However, what can be used to compare the two views is the relative change in the Top-FOV of a viewer over time. This change should correlate with the change in the content of the egocentric video. Intuitively, if a viewer is looking straight ahead while walking on a straight line, his Top-FOV is not going to have drastic changes. Therefore, we expect the viewer's egocentric view to have a stable visual content.\\

\noindent\textbf{Node Features:} We extract two unary features for each node, one captures the changes in the content covered by his FOV, and the other is the number of visible people in the content of the Top-FOV. 

To encode the relative change in the visual content of viewer $i$ visible in the top-view camera, we form the $T \times T$ matrix ($T$ denotes the number of frames in the top-view video) $U^{IOU}_{i}$ whose elements $U^{IOU}_{i}(f_p,f_q)$ indicate the IOU (intersection over union) of the Top-FOV of person $i$ in frames $f_p$ and $f_q$. For example, if the viewer's Top-FOV in frame 10 has high overlap with his FOV in frame 30 (thus $U^{IOU}_{i}(10,30)$ has a high value), we expect to see a high visual similarity between frames 10 and 30 in the egocentric video. Examples shown in the middle column of Figure \ref{fig:FOVvsGIST} (a). 

\begin{figure}
	\centering
	\includegraphics[width=1\linewidth]{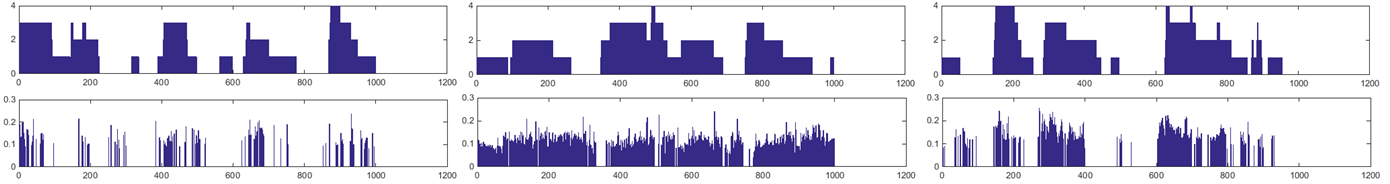}
	\caption{Examples of the 1D features capturing the number of visible humans time-series. Top row shows the number of visible people in each viewer's Top-FOV over time. Bottom row shows the summation of the detection scores at every frame in an egocentric video.}
	\label{fig:features_1D}
\end{figure}

Having the Top-FOV of viewer $i$ estimated, we then count the number of people within his Top-FOV at each time frame and store it in a ${1 \times T}$ vector $U^{n}_{i}$. To count the number of people, we used annotated bounding boxes. Figure \ref{fig:inFOVs} illustrates three viewers who have one human in their Top-FOV. Examples shown in the top row of figure \ref{fig:features_1D}.\\
\noindent\textbf{Edge Features:} Pairwise features are designed to capture the relationship among two different individuals. In the top-view videos, similar to the unary matrix $U^{IOU}_{i}$, we can form a $T \times T$ matrix $B^{IOU}_{ij}$ to describe the relationship between a pair of viewers (viewers/nodes $i$ and $j$), in which $B^{IOU}_{ij}(f_p,f_q)$ is defined as the intersection over union of the Top-FOVs of person $i$ in frame $f_p$ and person $j$ in frame $f_q$. Intuitively, if there is a high similarity between the Top-FOVs of person $i$ in frame $10$ and person $j$ in frame $30$, we would expect the 30th frame of viewer $j$'s egocentric video to be similar to the 10th frame of viewer $i$'s egocentric video. Two examples of such features are illustrated in the middle column of Figure \ref{fig:FOVvsGIST} (b). \\
 
\noindent \textbf{3.1.2 \ \ Modeling the Egocentric View Graph:} As in the top-view graph, we also construct a graph on the set of egocentric videos. Each node of this graph represents one egocentric video. Edges between the nodes capture the relationship between two egocentric videos.\\

\noindent\textbf{Node Features:} Similar to the top-view graph, each node is represented using two features. First, we compute pairwise similarity between GIST features \cite{GIST} of all video frames (for one viewer) and store the pairwise similarities in the matrix $U^{GIST}_{E_i}$, in which the element $U^{GIST}_{E_i}(f_p,f_q)$ is the GIST similarity between frame $f_p$ and $f_q$ of egocentric video $i$. Two examples of such features are illustrated in the left column of Figure \ref{fig:FOVvsGIST} (a). The GIST similarity is a function of the euclidean distance of the GIST feature vectors. 
\begin{equation}
U^{GIST}_{E_i}(f_1,f_2)=e^{-\gamma|g^{E_i}_{f_p}-g^{E_i}_{f_q}|}.%\frac{e^{-\gamma|g^{E_i}_{f_1}-g^{E_i}_{f_2}|}}{1+e^{-\gamma|g^{E_i}_{f_1}-g^{E_i}_{f_2}|}}
\end{equation}
In which $g^{E_i}_{f_p}$ and $g^{E_i}_{f_q}$ are the GIST descriptors of frame $f_p$ and $f_q$ of egocentric video $i$, and $\gamma$ is a constant which we empirically set to $0.5$.

The second feature is a time series counting the number of seen people in each frame. In order to have an estimate of the number of people, we run a pre-trained human detector using deformable part model~\cite{DPM2} on each egocentric frame. In order to make sure that our method is not including humans in far distances (which are not likely to be present in top-view), we exclude bounding boxes whose sizes are smaller than a certain threshold (determined considering an average human height of 1.7m and distance of the radius of the area being covered in the top view video.).
Each of the remaining bouding boxes, has a detection score (rescaled into the interval [0 1]) which has the notion of the probability of that bounding box containing a person. Scores of all detections in a frame are added and used as a count of people in that frame. Therefore, similar to the top-view feature, we can represent the node $E_i$ of egocentric video $i$ with a $1 \times T_{E_i}$ vector $U^{n}_{E_i}$ (Examples shown in the bottom row of figure \ref{fig:features_1D}.).\\
\noindent\textbf{Edge Features:} To capture the pairwise relationship between egocentric camera $i$ (containing $T_{E_i}$ frames) and egocentric camera $j$ (containing $T_{E_j}$ frames), we extract GIST features from all of the frames of both videos and form a $T_{E_i} \times T_{E_j}$ matrix $B^{GIST}_{ij}$ in which $B^{GIST}_{ij}(f_p,f_q)$ represents the GIST similarity between frame $f_p$ of video $i$ and frame $f_q$ of video $j$ (Examples shown in the left column of Figure \ref{fig:FOVvsGIST} (b).). 
\begin{equation}
B^{GIST}_{ij}(f_p,f_q)=e^{-\gamma|g^{E_i}_{f_p}-g^{E_j}_{f_q}|}.
\end{equation}

\subsection{Graph Matching}
The goal is to find a binary assignment matrix $x_{N^e \times N^t}$ ($N^e$ being the number of egocentric videos and $N^t$ being the number of people in the top-view video). $x(i,j)$ being 1 means egocentric video $i$ is matched to viewer $j$ in top-view. To capture the similarities between the elements of the two graphs, we define the affinity matrix $A_{N^eN^t \times N^eN^t}$ . $a_{ik,jl}$ is the affinity of edge $ij$ in the egocentric graph with edge $kl$ in the top-view graph. Reshaping matrix $x$ as a vector $x_{N^eN^t \times 1} \in \{0,1\}^{N^eN^t}$, the assignment problem is defined as the following:
\begin{equation}
\underset{x}{\operatorname{argmax}} \ x^TAx.
\end{equation} 
We compute $a_{ik,jl}$ based on the similarity between the feature descriptor of edge $ij$ in the egocentric graph $B^{GIST}_{ij}$ and the feature descriptor for edge $kl$ in the top-view graph $B^{IOU}_{kl}$. 

As described in the previous section, each of these features is a 2D matrix. $B^{GIST}_{ij}$ is a $T_{E_i} \times T_{E_j}$ matrix, $T_{E_i}$ and $T_{E_j}$ being the number of frames in egocentric videos $i$ and $j$, respectively. On the other hand, $B^{IOU}_{kl}$ is a $T_{t} \times T_{t}$ matrix, $T_{t}$ being the number of frames in the top-view video. $B^{GIST}_{ij}$ and $B^{IOU}_{kl}$ are not directly comparable due to two reasons. First, the two matrices are not of the same size (the videos do not necessarily have the same length). Second, the absolute time in the videos do not correspond to each other (videos are not time-synchronized). For example, the relationship between viewers $i$ and $j$ in the 100th frame of the top-view video does not correspond to frame number 100 of the egocentric videos. Instead, we expect to see a correlation between the GIST similarity of frame $100+d_i$ of egocentric video $i$ and frame $100+d_j$ of egocentric video $j$, and the intersection over union of in Top-FOVs of viewers $k$ and $l$ in frame 100. $d_i$ and $d_j$ are the time delays of egocentric videos $i$ and $j$ with respect to the top-view video.

\begin{figure}[t]
\begin{center}
		\begin{subfigure}[t]{0.48\textwidth}
			\includegraphics[width=1\linewidth]{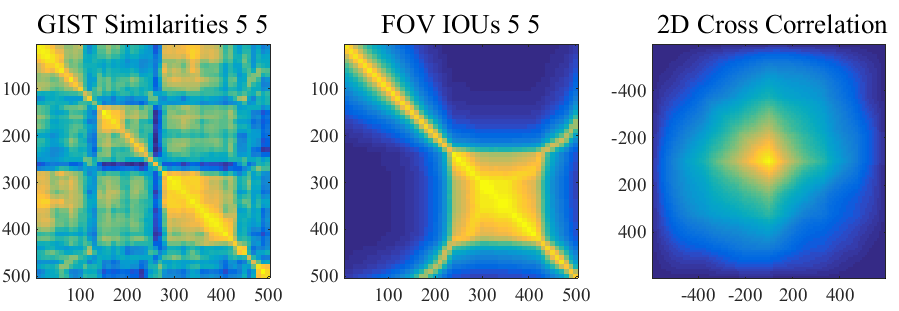}		
		\end{subfigure}\hspace{0.15 in} %\hskip 6pt
		\begin{subfigure}[t]{0.48\textwidth}
			\includegraphics[width=1\linewidth]{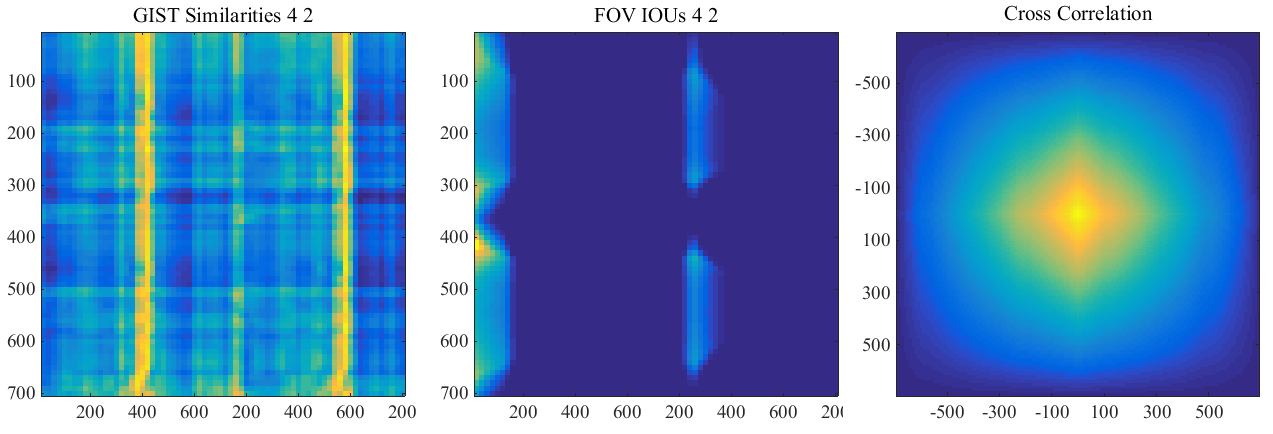}		
		\end{subfigure}
		
		\begin{subfigure}[t]{0.48\textwidth}
			\includegraphics[width=1\linewidth]{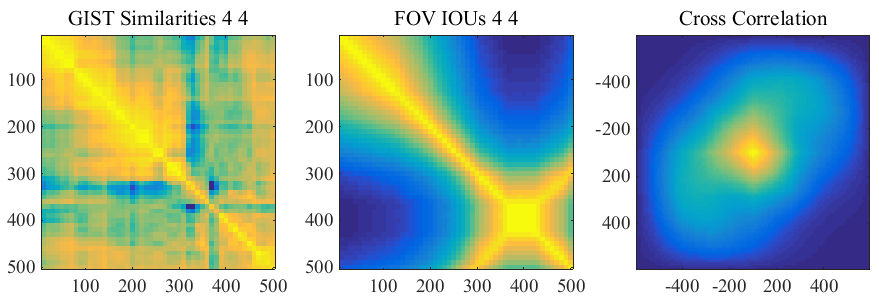}
			\caption{}	
		\end{subfigure}\hspace{0.15 in}
		\begin{subfigure}[t]{0.48\textwidth}
			\includegraphics[width=1\linewidth]{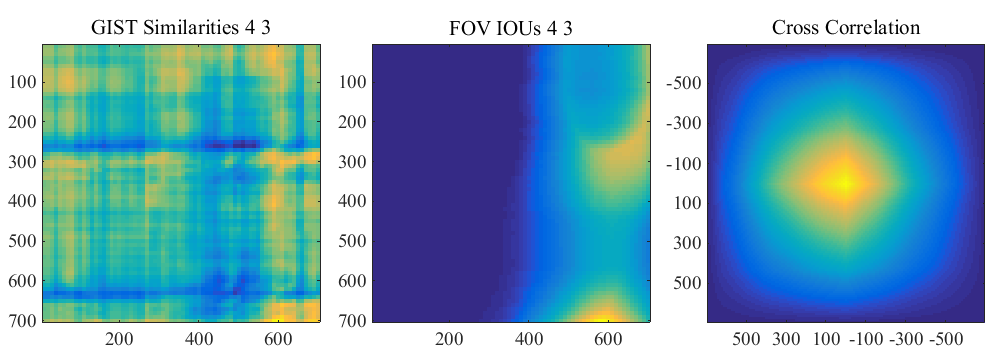}
			\caption{}	
		\end{subfigure}		
		\caption{(a) shows two different examples of the 2D features extracted from the \textbf{nodes} of the graphs (values color-coded). Left column shows the 2D matrices extracted from the pairwise similarities of the visual descriptors $U^{GIST}$, middle shows the 2D matrices computed by intersection over union of the FOV in the top-view camera $U^{IOU}$, and the rightmost column shows the result of the 2D cross correlation between the two. (b) shows the same concept, but between two \textbf{edges}. Again, the leftmost figure shows the pairwise similarity between GIST descriptors of one egocentric camera to another $B^{GIST}$. Middle, shows the pairwise intersection over union of the FOVs of the pair of viewers $B^{IOU}$, and the rightmost column illustrates their 2D cross correlation. The similarities between the GIST and FOV matrices capture the affinity of two nodes/edges in the two graphs.}
		\label{fig:FOVvsGIST}										
\end{center}
\end{figure}

To able to handle this misalignment, we define the affinity between the two 2D matrices as the maximum value of their 2D cross correlation. Hence, if egocentric videos $i$ and $j$ have $d_i$ and $d_j$ delays with respect to the top-view video, the cross correlation between $B^{GIST}_{ij}$ and $B^{IOU}_{kl}$ should be maximum when $B^{GIST}_{ij}$ is shifted $d_i$ units in the first, and $d_j$ units in the second dimension. 

\begin{equation}
A_{ikjl}=\text{max} (B^{GIST}_{ij} \ast B^{IOU}_{kl}).
\end{equation}

where $\ast$ denotes cross correlation. For the elements of $A$ for which $i=j$ and $k=l$, the affinity captures the compatibility of node $i$ in the egocentric graph, to node $k$ in the top-view graph. The compatibility between the two nodes is computed using 2D cross correlation between  
$U^{IOU}_{k}$ and $U^{GIST}_{E_i}$ and 1D cross correlation between $U^{n}_{k}$ and $U^{n}_{E_i}$. The overall compatibility of the two nodes is a weighted linear combination of the two:
\begin{equation}
\label{eq:node_similarity}
A_{ikik}=\alpha \text{max} (U^{GIST}_{E_i} \ast U^{IOU}_{k}) + (1-\alpha) \text{max}(U^{n}_{E_i} \ast U^{n}_{k}),
\end{equation}
where $\alpha$ is a constant between 0 and 1 specifying the contribution of each term. In our experiments, we set $\alpha$ to 0.9. Figure \ref{fig:FOVvsGIST} illustrates the features extracted from some of the nodes and edges in the two graphs.\\
%\begin{figure}
%	\centering
%	\includegraphics[width=1\linewidth]{Figures/GISTvsFOV/node_sim.png}
%	\caption{Edge Similarity.}
%	\label{fig:node_similarity}
%\end{figure}

\noindent\textbf{Soft Assignment} We employ the spectral graph matching method introduced in \cite{spectralMatching} to compute a soft assignment between the set of egocentric viewers and top-view viewers. In \cite{spectralMatching}, assuming that the affinity matrix is an empirical estimation of the pairwise assignment probability, and the assignment probabilities are statistically independent, $A$ is represented using it's rank one estimation which is computed by $\underset{p}{\operatorname{argmin}} \ |A-pp^T|$. In fact, the rank one estimation of $A$ is no different than it's leading eigenvector. Therefore, $p$ can be computed either using eigen decompositon, or estimated iteratively using power iteration. Considering vector $p$ as the assignment probablities, we can reshape $p_{N^eN^t \times 1}$ into a $N^e \times N^t$ soft assignment matrix $P$, for which after row normalization $P(i,j)$ represents the probability of matching egocentric viewer $i$ to viewer $j$ in the top-view video. \\

\noindent\textbf{Hard Assignment} Any soft to hard assignment method can be used to convert the soft assignment result (generated by spectral matching) to the hard binary assignment between the nodes of the graphs. We used the well known Munkres (also known as Hungarian) algorithm to obtain the final binary assignment.

\section{Experimental Results}
\label{sec:results}
In this section, we will mention details of our experimental setup and collected dataset, the measures we used to evaluate the performance of our method, and the performance of our proposed method alongside with some baselines.
\subsection{Dataset}
We collected a dataset containing 50 test cases of videos shot in different indoor and outdoor conditions. Each test case, contains one top-view video and several egocentric videos captured by the people visible in the top-view camera. Depending on the subset of egocentric cameras that we include, we can generate up to 2,862 instances of our assignment problem (will be explained in more detail in section 4.2.4). Overall, our dataset contains more than 225,000 frames. Number of people visible in the top-view cameras varies from 3 to 10, number of egocentric cameras varies from 1 to 6, and the ratio of number of available egocentric cameras to the number of visible people in the top-view camera varies from  0.16 to 1. Lengths of the videos vary from 320 frames (10.6 seconds) up to 3132 frames (110 seconds).
\subsection{Evaluation}	
We evaluate our method in terms of answering the two questions we asked. First, given a top-view video and a set of egocentric videos, can we verify if the top-view video is capturing the egocentric viewers? We analyze the capability of our method in answering this question in section 4.2.1. 

Second, knowing that a top-view video contains the viewers recording a set of egocentric videos, can we determine which viewer has recorded which video? We answer this question in sections 4.2.2 and 4.2.3.
\subsubsection{4.2.1. Ranking Top-view Videos:} 
\label{sec:sceneRanking}
We design an experiment to evaluate if our graph matching score is a good measure for the similarity between the set of egocentric videos and a top-view video. Having a set of egocentric videos from the same test case (recorded in the same environment), and 50 different top-view videos (from different test cases), we compare the similarity of each of the top-view graphs to the egocentric graph. After computing the hard assignment for each top view video(resulting in the assignment vector $x$), the score $x^TAx$ is associated to that top-view video. This score is effectively the summation of all the similarities between the corresponding nodes and edges of the two graphs. Using this score rank all the top-view videos. The ranking accuracy is measured by measuring the rank of the ground truth top-view video, and computing the cumulative matching curves shown in figure \ref{fig:all_results}(a). The blue curve shows the ranking accuracy when we compute the scores only based on the unary features. The red curve shows the ranking accuracy when we consider both the unary and pairwise features for performing graph matching. The dashed black line shows the accuracy of randomly ranking the top-view videos. It can be observed that both the blue and red curves outperform the random ranking. This shows that our graph matching score is a meaningful measure for estimating the similarity between the two graphs. In addition, the red curve, outperforming the blue curve shows the effectiveness of our pairwise features. In general, this experiment answers the first question. We can in fact use the graph matching score as a cue for narrowing down the search space among the top-view videos, for finding the one corresponding to our set of the egocentric cameras.
\subsubsection{4.2.2. Viewer Ranking Accuracy:}
We evaluate our soft assignment results, in terms of ranking capability. In other words, we can look at our soft assignment as a measure to sort the viewers in the top-view video based on their assignment probability to each egocentric video. Computing the ranks of the correct matches, we can plot the cumulative matching curves to illustrate their performance. %We also compute the area under curve for having a quantitative measurement of our framework. 

We compare our method with three baselines in figure \ref{fig:all_results} (b). First, random ranking (dashed black line), in which for each egocentric video we randomly rank the viewers present in the top-view video. Second, sorting the top-view viewers based on the similarities of their 1D unary features to the 1D unary features of each egocentric camera (i.e., number of visible humans illustrated by the blue curve). Third, sorting the top-view viewers based on their 2D unary feature (GIST vs. FOV, shown by the green curve). Note that here (the blue and green curves), we are ignoring the pairwise relationships (edges) in the graphs. The consistent improvement of our method (red curve) over the baselines, justifies the effectiveness of our representation, and shows the contribution of each stage. 

\subsubsection{4.2.3. Assignment Accuracy:} 
\begin{figure}[t]
	\begin{center}
		\begin{subfigure}[t]{0.32\textwidth}
			\includegraphics[width=\textwidth]{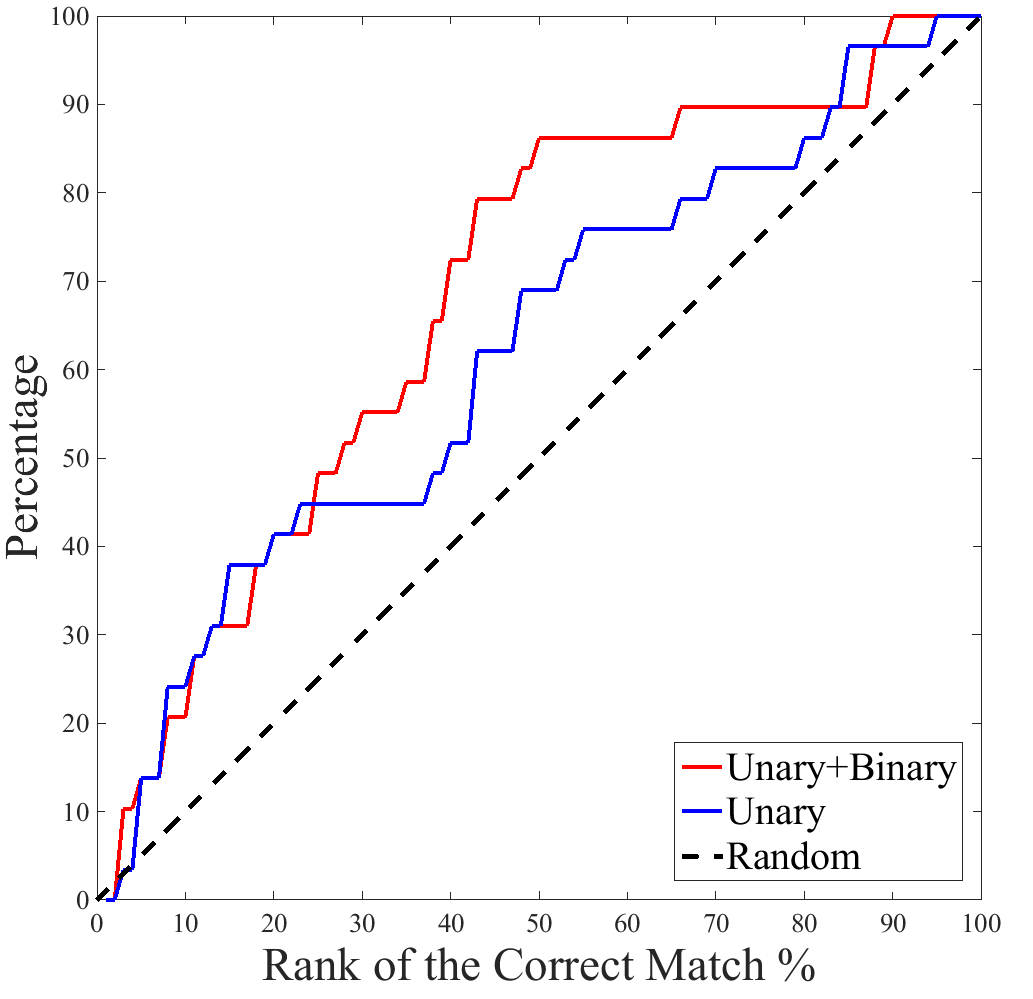}
			\caption{}
			\label{fig:sceneMatching_teaser}			
		\end{subfigure}\hfill	
		\begin{subfigure}[t]{0.32\textwidth}
			\includegraphics[width=\textwidth]{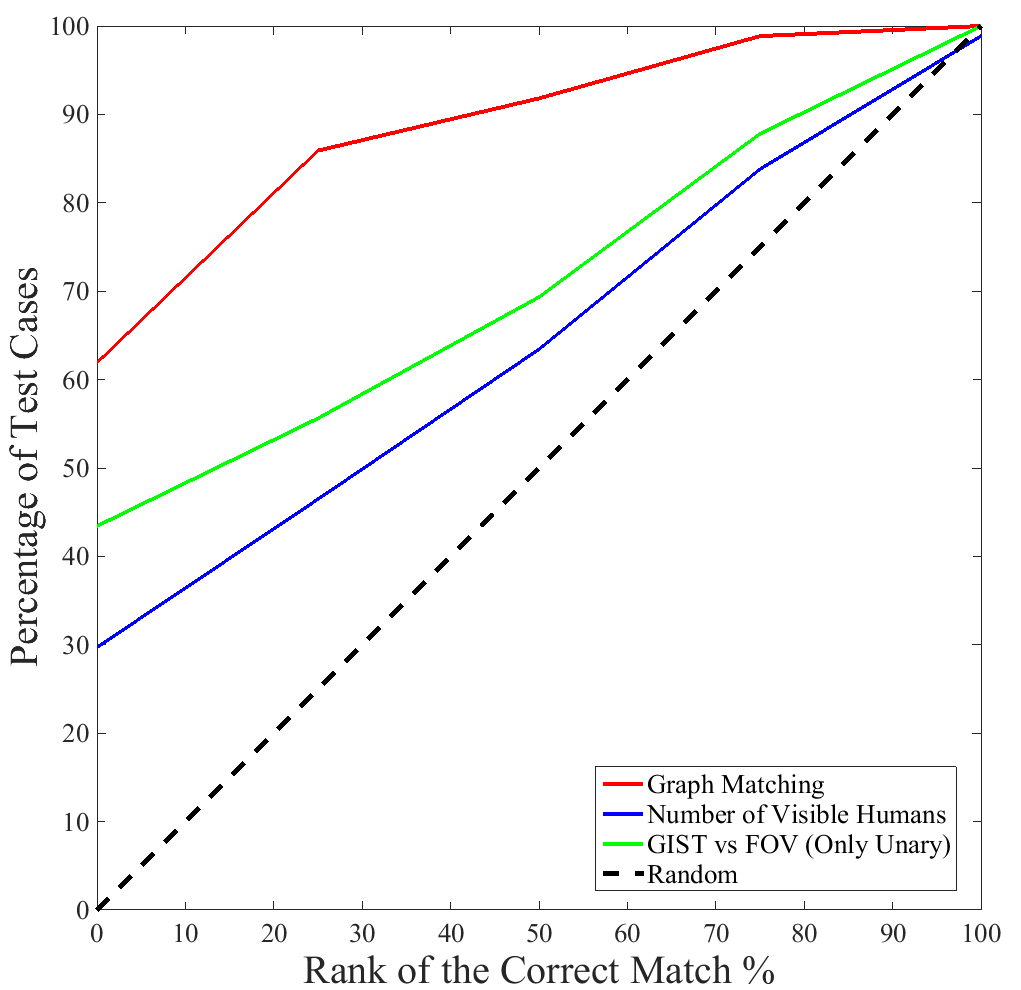}
			\caption{}
			\label{fig:fullSet_cmc}			
		\end{subfigure}%\hfill	
		\begin{subfigure}[t]{0.34\textwidth}
			\includegraphics[width=\textwidth]{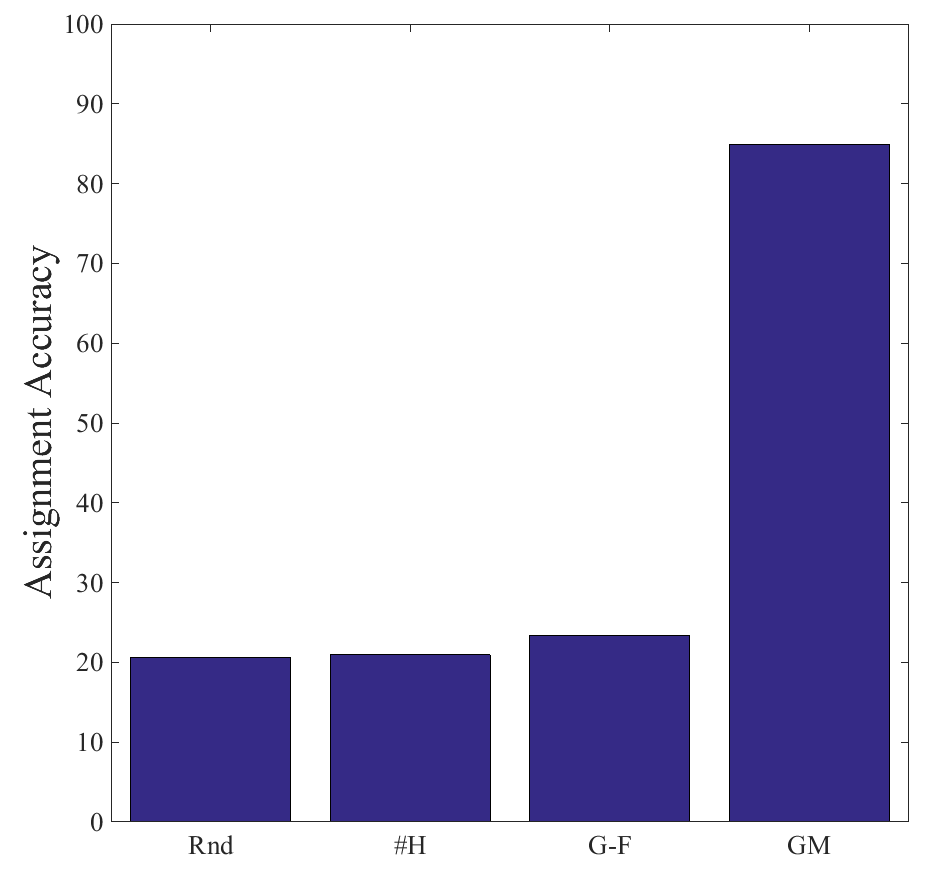}
			\caption{}
			\label{fig:fullSet_bar}			
		\end{subfigure}%\hfill			
		\caption{(a) shows the cumulative matching curve for ranking top-view videos. The blue curve shows the accuracy achieved only using the node similarities. Red is the accuracy considering both node and edge similarities in the graph matching. (b) shows the cumulative matching curve for ranking the viewers in the top-view video. The red, green and blue curves belong to ranking based on spectral graph matching scores, cross correlation between only the 2D, and only the 1D unary scores, respectively. The dashed black line shows random ranking accuracy (c) shows the assignment accuracy based on random assignment, using the number of humans, using unary features, and using spectral graph matching.}	
		\label{fig:all_results}			
	\end{center}
\end{figure}

In order to answer the second question, we need to evaluate the accuracy of our method in terms of node assignment accuracy. Having a set of egocentric videos and a top-view video containing the egocentric viewers, we evaluate the percentage of the egocentric videos which were correctly matched to their corresponding viewer. We evaluate the hard-assignment accuracy of our method and compare it with three baselines in figure \ref{fig:all_results}(c). First, random assignment (Rnd), in which we randomly assign each egocentric video to one of the visible viewers in the top-view video. Second, Hungarian bipartite matching only on the 1D unary features denoted as H. Third, Hungarian bipartite matching only on the 2D unary feature (GIST vs. FOV, denoted as G-F), ignoring the pairwise relationships (edges) in the graphs. 

The consistent improvement of our method using both unary and pairwise features in graph matching (denoted as GM) over the baselines shows the significant contribution of pairwise features in the assignment accuracy. As a result, the promising accuracy acquired by graph matching answers the second question. Knowing a top-view camera is capturing a set of egocentric viewers, we can use visual cues in the egocentric videos and the top-view video,
to decide which viewer is capturing which egocentric video.

%[[fix figure 7. also use legend in the bar chart and title to each panel]]
%\setlength{\tabcolsep}{4pt}
%\begin{table}
%	\begin{center}
%		\caption{Assignment Accuracy}
%		\label{table:headings}
%		\begin{tabular}{llll}
%			\hline\noalign{\smallskip}
%			Percentage of Present IDs & Unary Scores & GIST vs FOV & fusion \\
%			\noalign{\smallskip}
%			\hline
%			\noalign{\smallskip}
%			100 & 64  & 100 & 100\\
%			80 & 48.3 & 33.3 & 53.3\\
%			60 & 56.6 & 15 & 30\\
%			40 & - & - & 20\\
%			\hline
%		\end{tabular}
%	\end{center}
%\end{table}
%\setlength{\tabcolsep}{1.4pt}

\subsubsection{4.2.4. Effect of Number of Egocentric Cameras:} In sections 4.2.2 and 4.2.3, we evaluated the performance of our method given all the available egocentric videos present in each set as the input to our method. In this experiment, we compare the accuracy of our assignment and ranking framework as a function of the completeness ratio ($\frac{n_{Ego}}{n_{Top}}$) of our egocentric set. Each of our sets contain $3<N^t<11$ viewers in the top-view camera, and $2<N^e<8$ egocentric videos. We evaluated the accuracy of our method and baselines when using different subsets of the egocentric videos. 
A total of $2^{N^e}-1$ non-empty subsets of egocentric videos is possible 
depending on which egocentric video out of $N^e$ are included (all possible non-empty subsets). We evaluate our method on each subset separately. 

\begin{figure}
	\begin{center}
		\begin{subfigure}[t]{0.24\textwidth}
			\includegraphics[width=\textwidth]{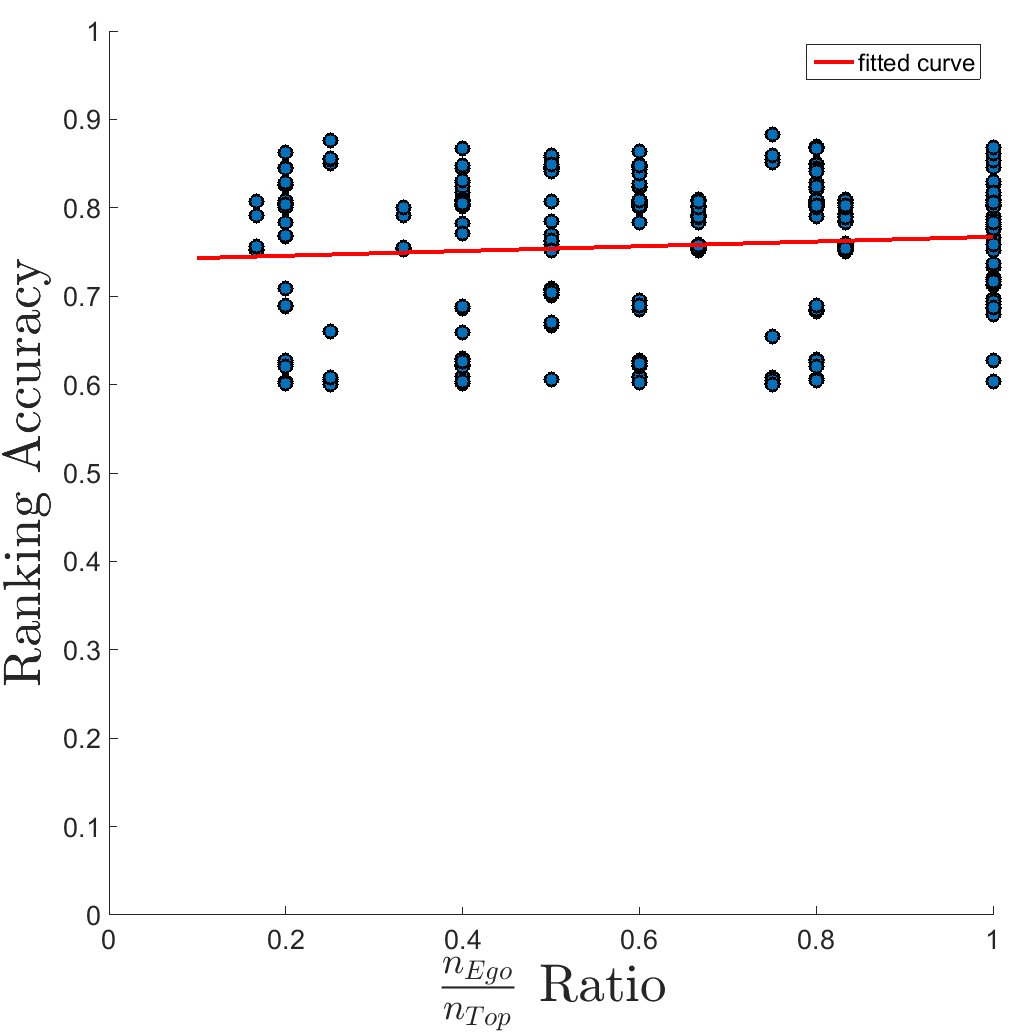}	
			\caption{}			
			\label{fig:CMC_car_overall}		
		\end{subfigure}\hfill	
				\begin{subfigure}[t]{0.24\textwidth}
					\includegraphics[width=\textwidth]{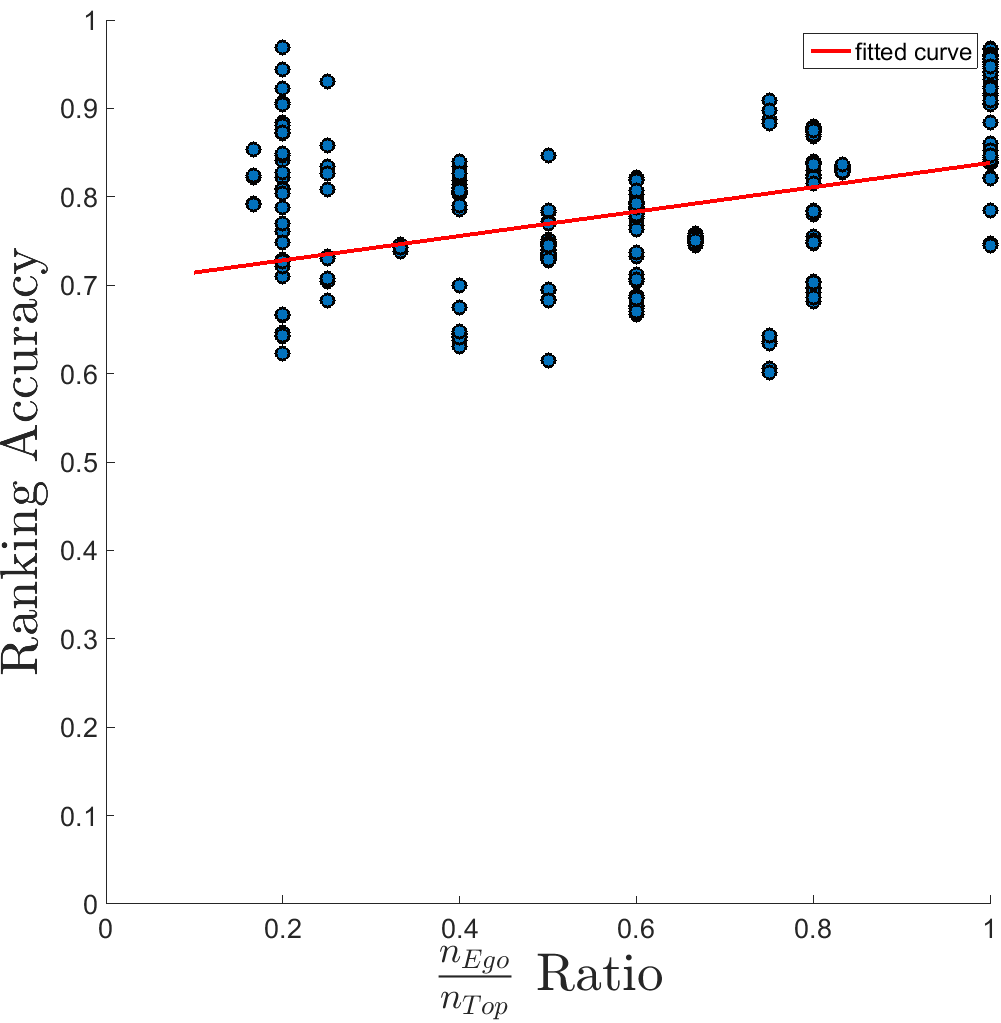}
					\caption{}
					\label{fig:CMC_car_overall}			
				\end{subfigure}\hfill
		\begin{subfigure}[t]{0.24\textwidth}
			\includegraphics[width=\textwidth]{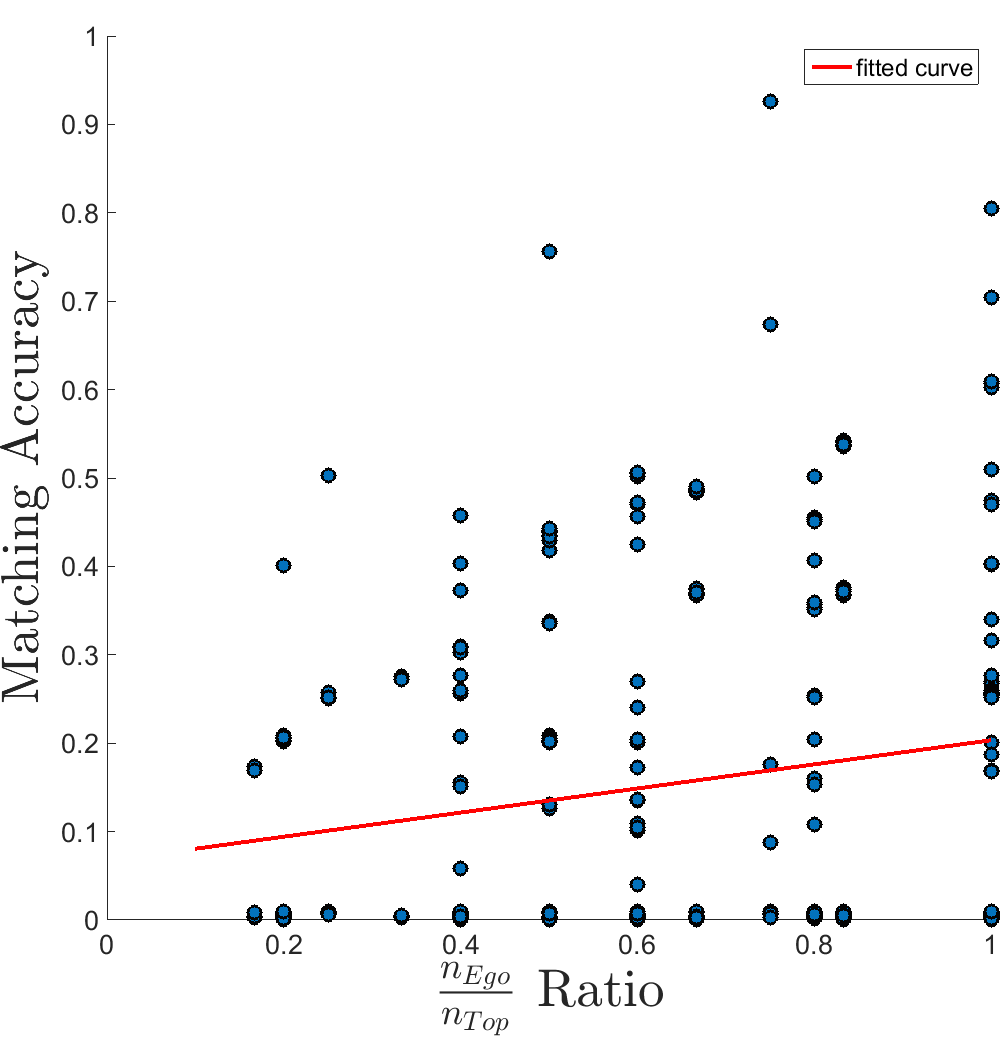}	
			\caption{}		
			\label{fig:CMC_car_overall}			
		\end{subfigure}\hfill			
		\begin{subfigure}[t]{0.24\textwidth}
				\includegraphics[width=\textwidth]{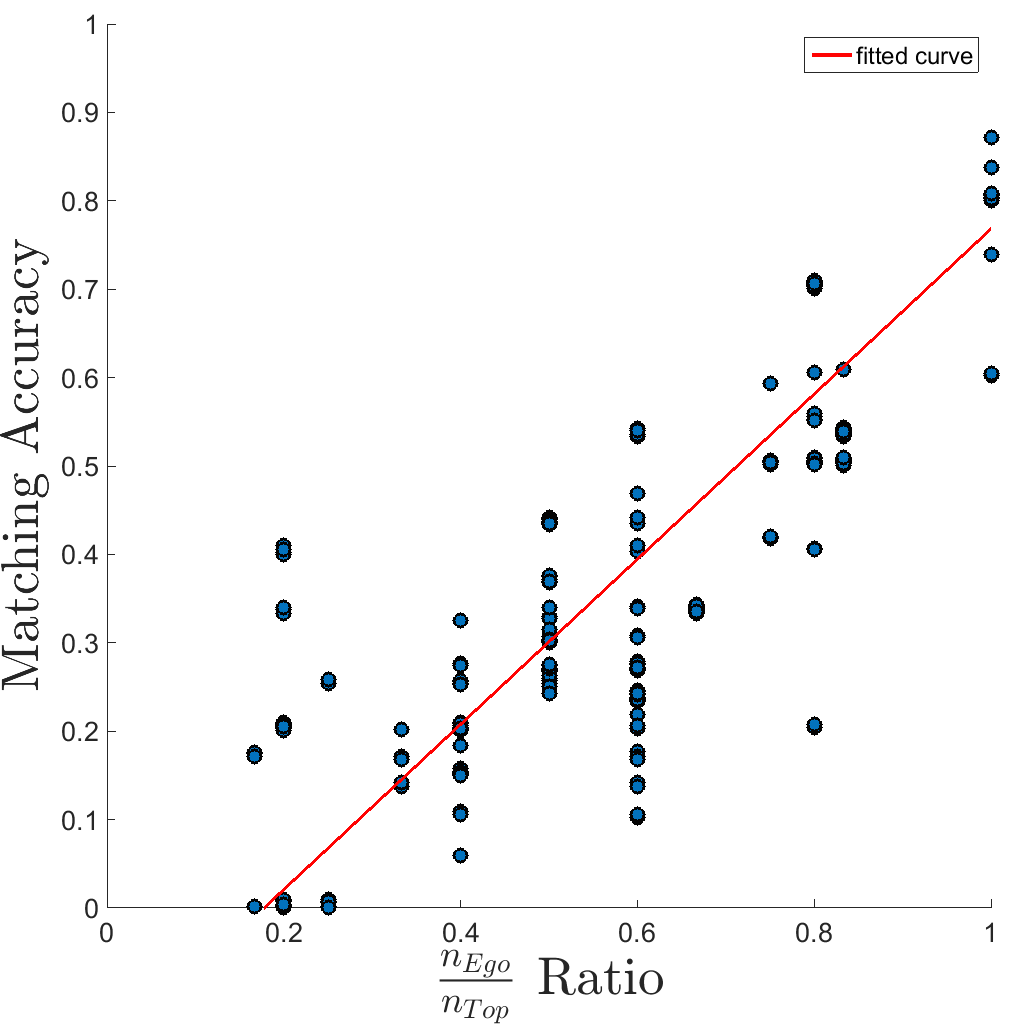}			\caption{}	
				\label{fig:CMC_car_overall}			
		\end{subfigure}\hfill
				\caption{Effect of the relative number of egocentric cameras referred to as completeness ratio ($\frac{n_{Ego}}{n_{Top}}$). (a) shows the ranking accuracy vs $\frac{n_{Ego}}{n_{Top}}$, only using the unary features. (b) shows the same evaluation using the graph matching output. (c) shows the accuracy of the hard assignment computed based on Hungarian bipartite matching on top of the unary features, and (d) shows the hard-assignment computed based on the spectral graph matching.}
				\label{fig:Ego2TopRatio}			
	\end{center}
\end{figure}

Figure \ref{fig:Ego2TopRatio} illustrates the assignment and ranking accuracies versus the ratio of the available egocentric videos to the number of visible people in the top-view camera. It shows that as the completeness ratio increases, the assignment accuracy drastically improves. Intuitively, having more egocentric cameras gives more information about the structure of the graph (by providing more pairwise terms) which leads to improvement in the spectral graph matching and assignment accuracy. 

%\subsubsection{False Positives:} 

%\subsubsection{Effect of Video Length}
%Due to the fact that our similarity measures in the spectral graph matching step are based on feature vectors whose size increase with the length of the video (in terms of time), we expect to observe more discriminative power for longer sequences. In order to verify, we performed matching and ranking using different number of frames. The results are illustrated in figure. It can be observed that as the length of the videos increase, the discriminative power of our graph matching scheme increases. 
%\subsubsection{Robustness to Synthetic Noise}

%\section{Failure Cases}
\section{Conclusion and Discussion}
\label{sec:conclusion}
In this work, we studied the problems of matching and assignment between a set of egocentric cameras and a top view video. Our experiments suggest that capturing the pattern of change in the content of the egocentric videos, along with capturing the relationships among them can help to identify the viewers in top-view. To do so, we utilized a spectral graph matching technique. We showed that the graph matching score, is a meaningful criteria for narrowing down the search space in a set of top-view videos. Further, the assignment found by our framework is capable of associating egocentric videos to the viewers in the top-view camera. We conclude that meaningful features can be extracted from single, and pairs of egocentric camera(s) and incorporating the temporal information of the video(s).

%\begin{figure}
%	\centering
%	\includegraphics[width=1\linewidth]{Figures/GISTvsFOV/edge_sim.png}
%	\caption{Edge Similarity.}
%	\label{fig:example}
%\end{figure}

%\begin{figure}
%	\centering
%	\includegraphics[width=1\linewidth]{Figures/CaseByCase.png}
%	\caption{Evaluating matching as a ranking problem.}
%	\label{fig:example}
%\end{figure}
%
%\begin{figure}
%	\centering
%	\includegraphics[width=1\linewidth]{Figures/Bars/CaseByCase_ID_end_1_2_3.png}
%	\caption{Evaluating matching as a ranking problem.}
%	\label{fig:example}
%\end{figure}
%
%\begin{figure}
%	\centering
%	\includegraphics[width=1\linewidth]{Figures/Bars/CaseByCase_ID_end_1_2.png}
%	\caption{Evaluating matching as a ranking problem.}
%	\label{fig:example}
%\end{figure}
%
%\begin{figure}
%	\centering
%	\includegraphics[height=6.5cm]{Figures/cmc/all.png}
%	\caption{Evaluating matching as a ranking problem.}
%	\label{fig:example}
%\end{figure}
%
%\begin{figure}
%	\centering
%	\includegraphics[height=6.5cm]{Figures/cmc/all_last2.png}
%	\caption{Evaluating matching as a ranking problem.}
%	\label{fig:example}
%\end{figure}
%
%\begin{figure}
%	\centering
%	\includegraphics[height=6.5cm]{Figures/cmc/all_last3.png}
%	\caption{Evaluating matching as a ranking problem.}
%	\label{fig:example}
%\end{figure}

\clearpage

\bibliographystyle{splncs}
\bibliography{Thesis_bib}

\begin{thebibliography}{10}

\bibitem{ardeshiregocentric}
Ardeshir, S., Borji, A.:
\newblock From egocentric to top-view

\bibitem{egoActionFathi}
Fathi~A, Farhadi~A, R.J.:
\newblock Understanding egocentric activities.
\newblock Computer Vision (ICCV), 2011 IEEE International Conference on. IEEE
  (2011)

\bibitem{egoDailyAction}
Fathi~A, Li~Y, R.J.:
\newblock Learning to recognize daily actions using gaze.
\newblock Computer Vision–ECCV (2012)

\bibitem{egoFOVLocalization}
Bettadapura, Vinay, I.E., Pantofaru., C.:
\newblock Egocentric field-of-view localization using first-person
  point-of-view devices.
\newblock Applications of Computer Vision (WACV), IEEE Winter Conference on.
  (2015)

\bibitem{spectralMatching}
Egozi, Amir, Y.K., Guterman., H.:
\newblock A probabilistic approach to spectral graph matching.
\newblock Pattern Analysis and Machine Intelligence, IEEE Transactions on
  (2013)

\bibitem{theWayTheyMove}
Dicle, Caglayan, O.C., Sznaier., M.:
\newblock The way they move: Tracking multiple targets with similar appearance.
\newblock Proceedings of the IEEE International Conference on Computer Vision
  (2013)

\bibitem{egoKanade}
Kanade, T., Hebert., M.:
\newblock First-person vision.
\newblock Proceedings of the IEEE 100.8 (2012)

\bibitem{egoEvolutionSurvey}
Betancourt~A, Morerio~P, R.C.R.M.:
\newblock The evolution of first person vision methods: A survey.
\newblock Circuits and Systems for Video Technology, IEEE Transactions on
  (2015)

\bibitem{egoObjectDetection}
Fathi, Alireza, X.R., Rehg., J.M.:
\newblock Learning to recognize objects in egocentric activities.
\newblock Computer Vision and Pattern Recognition (CVPR), 2011 IEEE Conference
  On (2011)

\bibitem{egoVideoSummarization}
Lu, Z., Grauman., K.:
\newblock Story-driven summarization for egocentric video.
\newblock Computer Vision and Pattern Recognition (CVPR), IEEE Conference On
  (2013)

\bibitem{egoli2013learning}
Li, Y., Fathi, A., Rehg, J.:
\newblock Learning to predict gaze in egocentric video.
\newblock In: Proceedings of the IEEE International Conference on Computer
  Vision. (2013)  3216--3223

\bibitem{egoPolatsekNovelty}
Polatsek, P., Benesova, W., Paletta, L., Perko, R.:
\newblock Novelty-based spatiotemporal saliency detection for prediction of
  gaze in egocentric video

\bibitem{Borji2014look}
Borji, A., Sihite, D.N., Itti, L.:
\newblock What/where to look next? modeling top-down visual attention in
  complex interactive environments.
\newblock Systems, Man, and Cybernetics: Systems, IEEE Transactions on
  \textbf{44}(5) (2014)  523--538

\bibitem{egoMobileFixedObjectDetection}
Alahi, Alexandre, M.B., Kunt., M.:
\newblock Object detection and matching with mobile cameras collaborating with
  fixed cameras.
\newblock Workshop on Multi-camera and Multi-modal Sensor Fusion Algorithms and
  Applications-M2SFA2 (2008)

\bibitem{egoMobileFixedMasterSlave}
Alahi~A, Marimon~D, B.M.K.M.:
\newblock A master-slave approach for object detection and matching with fixed
  and mobile cameras.
\newblock InImage Processing, 2008. ICIP 2008. 15th IEEE International
  Conference (2008)

\bibitem{egoExo}
Ferland~F, Pomerleau~F, L.D.C.M.F.:
\newblock Egocentric and exocentric teleoperation interface using real-time, 3d
  video projection.
\newblock InHuman-Robot Interaction (HRI), 2009 4th ACM/IEEE International
  Conference on (2009)

\bibitem{egoPredictingGaze}
Park, Hyun, E.J., Sheikh., Y.:
\newblock Predicting primary gaze behavior using social saliency fields.
\newblock Proceedings of the IEEE International Conference on Computer Vision.
  (2013)

\bibitem{egoWisdomOfTheCrowd}
Hoshen, Yedid, G.B.A., Peleg., S.:
\newblock Wisdom of the crowd in egocentric video curation.
\newblock Proceedings of the IEEE Conference on Computer Vision and Pattern
  Recognition Workshops. 2014 (2014)

\bibitem{egoSocialInteractions}
Fathi, Alireza, J.K.H., Rehg., J.M.:
\newblock Social interactions: A first-person perspective.
\newblock Computer Vision and Pattern Recognition (CVPR), IEEE Conference on.
  (2012)

\bibitem{egoMultiTaskClustering}
Yan, Yan, e.a.:
\newblock Egocentric daily activity recognition via multitask clustering.
\newblock Image Processing, IEEE Transactions on (2015)

\bibitem{egoYouDoILearn}
Damen~D, Leelasawassuk~T, H.O.C.A.M.C.W.:
\newblock You-do, i-learn: Discovering task relevant objects and their modes of
  interaction from multi-user egocentric video.
\newblock BMVC (2014)

\bibitem{reidCPS}
Cheng~DS, Cristani~M, S.M.B.L.M.V.:
\newblock Custom pictorial structures for re-identification.
\newblock BMVC (2011)

\bibitem{reidReimannian}
Bak~S, Corvee~E, B.F.T.M.:
\newblock Multiple-shot human re-identification by mean riemannian covariance
  grid.
\newblock InAdvanced Video and Signal-Based Surveillance (AVSS), 8th IEEE
  International Conference on (2011)

\bibitem{reidSDALF}
Bazzani~L, Cristani~M, M.V.:
\newblock Symmetry-driven accumulation of local features for human
  characterization and re-identification.
\newblock omputer Vision and Image Understanding. (2013)

\bibitem{egoHeadMotion}
Cheng~DS, Cristani~M, S.M.B.L.M.V.:
\newblock Head motion signatures from egocentric videos.
\newblock InComputer Vision--ACCV. Springer International Publishing. (2014)

\bibitem{egoSurfing}
Yonetani, Ryo, K.M.K., Sato., Y.:
\newblock Ego-surfing first person videos.
\newblock Computer Vision and Pattern Recognition (CVPR), 2015 IEEE Conference
  on. IEEE, (2015)

\bibitem{TagrefinementCVPR14}
Zamir, A.R., Ardeshir, S., Shah, M.:
\newblock Gps-tag refinement using random walks with an adaptive damping
  factor.
\newblock In: 2014 IEEE Conference on Computer Vision and Pattern Recognition,
  IEEE (2014)  4280--4287

\bibitem{egoWhereAmI}
Kiefer, Peter, I.G., Raubal., M.:
\newblock Where am i? investigating map matching during self‐localization
  with mobile eye tracking in an urban environment.
\newblock Transactions in GIS 18.5 (2014)

\bibitem{GISobject}
Shervin~Ardeshir, Amir Roshan~Zamir, A.T., Shah., M.:
\newblock Gis-assisted object detection and geospatial localization.
\newblock In European Conference on Computer Vision. ECCV (2014)  602--617

\bibitem{Geosemantic}
Ardeshir, S., Malcolm Collins-Sibley, K., Shah, M.:
\newblock Geo-semantic segmentation.
\newblock In: Proceedings of the IEEE Conference on Computer Vision and Pattern
  Recognition. (2015)  2792--2799

\bibitem{GIST}
Torralba, A.:
\newblock Contextual priming for object detection.
\newblock In: International Journal of Computer Vision, Vol. 53(2), 169-191.
  (2003)

\bibitem{DPM2}
Felzenszwalb, P.F., Girshick, R.B., McAllester, D., Ramanan, D.:
\newblock Object detection with discriminatively trained part based models.
\newblock IEEE Transactions on Pattern Analysis and Machine Intelligence
  \textbf{32}(9) (2010)  1627--1645

\end{thebibliography}



@misc{SURF,
	author = { Bay, Herbert, Tinne Tuytelaars, and Luc Van Gool.},
	title = {Surf: Speeded up robust features.},
	note = {In Computer vision–ECCV 2006, pp. 404-417. Springer Berlin Heidelberg.},
	year = 2006
}

 

@misc{KLT,
	author = {Tomasi, Carlo, and Takeo Kanade.},
	title = {Detection and tracking of point features.},
	note = { Pittsburgh: School of Computer Science, Carnegie Mellon Univ.},
	year = 1991
}




@misc{TroppCSS,
	author = {Tropp, Joel A.},
	title = {Column subset selection, matrix factorization, and eigenvalue optimization.},
	note = {In Proceedings of the Twentieth Annual ACM-SIAM Symposium on Discrete Algorithms, pp. 978-986. Society for Industrial and Applied Mathematics.},
	year = 2009
}




@misc{CenRao2,
	author = {Rao, Cen, Alexei Gritai, Mubarak Shah, and Tanveer Syeda-Mahmood.},
	title = {View-invariant alignment and matching of video sequences.},
	note = {In Computer Vision, 2003. Proceedings. Ninth IEEE International Conference on, pp. 939-945. IEEE, 2003.},
	year = 2003
}




@misc{CenRao,
	author = {Rao, Cen, Alper Yilmaz, and Mubarak Shah. },
	title = {View-invariant representation and recognition of actions.},
	note = {International Journal of Computer Vision 50, no. 2 (2002): 203-226.},
	year = 2002
}



@misc{BrubakerLOST,
	author = {Brubaker, Marcus, Andreas Geiger, and Raquel Urtasun.},
	title = {Lost! leveraging the crowd for probabilistic visual self-localization.},
	note = {In Computer Vision and Pattern Recognition (CVPR), 2013 IEEE Conference on, pp. 3057-3064. IEEE,},
	year = 2013
}



@misc{ReillyDetectionAndTracking,
	author = {Reilly, Vladimir, Haroon Idrees, and Mubarak Shah.},
	title = {Detection and tracking of large number of targets in wide area surveillance.},
	note = {In Computer Vision–ECCV 2010, pp. 186-199. Springer Berlin Heidelberg.},
	year = 2010
}


@misc{KroegerSFM,
	author = {Kroeger, Till, and Luc Van Gool.},
	title = {Video registration to SfM models.},
	note = {In Computer Vision–ECCV 2014, pp. 1-16. Springer International Publishing,},
	year = 2014
}



@misc{YuMotionPatterns,
	author = {Yu, Qian, and Gérard Medioni.},
	title = {Motion pattern interpretation and detection for tracking moving vehicles in airborne video.},
	note = {In Computer Vision and Pattern Recognition, 2009. CVPR 2009. IEEE Conference on, pp. 2671-2678. IEEE,},
	year = 2009
}



@misc{XiaoVehicleDetectionAndTracking,
	author = {Xiao, Jiangjian, Hui Cheng, Harpreet Sawhney, and Feng Han.},
	title = {Vehicle detection and tracking in wide field-of-view aerial video.},
	note = {In Computer Vision and Pattern Recognition (CVPR), 2010 IEEE Conference on, pp. 679-684. IEEE,},
	year = 2010
}




@misc{IrscharaSFM,
	author = {Irschara, Arnold, Christopher Zach, Jan-Michael Frahm, and Horst Bischof.},
	title = {From structure-from-motion point clouds to fast location recognition.},
	note = {In Computer Vision and Pattern Recognition, 2009. CVPR 2009. IEEE Conference on, pp. 2599-2606. IEEE,},
	year = 2009
}




@misc{XiaoGeo,
	author = {Xiao, Jiangjian, Hui Cheng, Feng Han, and Harpreet Sawhney.},
	title = {Geo-spatial aerial video processing for scene understanding and object tracking.},
	note = {In Computer Vision and Pattern Recognition, 2008. CVPR 2008. IEEE Conference on},
	year = 2008
}

@misc{KumarRegistration,
	author = {Kumar, Rakesh, Harpreet S. Sawhney, Jane C. Asmuth, Art Pope, and Steve Hsu.},
	title = {Registration of video to geo-referenced imagery.},
	note = {In Pattern Recognition, 1998. Proceedings. Fourteenth International Conference on, vol. 2, pp. 1393-1400. IEEE,},
	year = 1998
}



@misc{NawazTrajectory,
	author = {Nawaz, Tasin, Andrea Cavallaro, and Bernhard Rinner.},
	title = {Trajectory clustering for motion pattern extraction in aerial videos.},
	note = {In Image Processing (ICIP), 2014 IEEE International Conference on},
	year = 2014
}

@misc{WPAFB,
	author = {},
	title = {},
	note = {https://www.sdms.afrl.af.mil/index.php?collection=wpafb2009},
	year = 
}



@misc{DeshpandeCSS,
	author = {Deshpande, Amit, and Luis Rademacher.},
	title = {Efficient volume sampling for row/column subset selection.},
	note = {Foundations of Computer Science (FOCS), 2010 51st Annual IEEE Symposium on. IEEE.},
	year = 2010
}

@misc{DeshpandeCSS,
	author = {Wei, Hua-Liang, and Stephen Billings.},
	title = {Feature subset selection and ranking for data dimensionality reduction.},
	note = {Pattern Analysis and Machine Intelligence, IEEE Transactions on 29.1: 162-166.},
	year = 2007
}
 


@misc{TroppCSS,
	author = {Tropp, Joel A.},
	title = {Column subset selection, matrix factorization, and eigenvalue optimization.},
	note = {Proceedings of the Twentieth Annual ACM-SIAM Symposium on Discrete Algorithms. Society for Industrial and Applied Mathematics.},
	year = 2009
}


@misc{ZhangTrajectoryClustering,
	author = {Zhang, Zhang, Kaiqi Huang, and Tieniu Tan.},
	title = {Comparison of similarity measures for trajectory clustering in outdoor surveillance scenes.},
	note = {Pattern Recognition, 2006. ICPR 2006. 18th International Conference on. Vol. 3. IEEE.},
	year = 2006
}

@misc{LeeTrajectoryClustering,
	author = {Lee, Jae-Gil, Jiawei Han, and Kyu-Young Whang.},
	title = {Trajectory clustering: a partition-and-group framework.},
	note = {Proceedings of the 2007 ACM SIGMOD international conference on Management of data. ACM.},
	year = 2007
}


@misc{LeeTrajectoryMatching,
	author = {Lee, HyungJune, et al.},
	title = {Localization of mobile users using trajectory matching.},
	note = {Proceedings of the first ACM international workshop on Mobile entity localization and tracking in GPS-less environments. ACM.},
	year = 2008
}


@misc{CroitoruTrajectoryMatching,
	author = {Croitoru, Arie, Peggy Agouris, and Anthony Stefanidis.},
	title = {3D trajectory matching by pose normalization.},
	note = {Proceedings of the 13th annual ACM international workshop on Geographic information systems. ACM.},
	year = 2005
}



@misc{HsiehTrajectoryMatching,
	author = {Hsieh, Jun-Wei, Shang-Li Yu, and Yung-Sheng Chen.},
	title = {Motion-based video retrieval by trajectory matching.},
	note = {Circuits and Systems for Video Technology, IEEE Transactions on 16.3 (2006): 396-409.},
	year = 2006
}


@misc{Geosemantic,
	author = {Shervin Ardeshir, Kofi Malcolm Collins-Sibley, and Mubarak Shah.},
	title = {Geo-semantic Segmentation.},
	note = {Proceedings of the IEEE Conference on Computer Vision and Pattern Recognition.},
	year = 2015
}


@article{GISobject,
	title = "Gis-assisted object detection and geospatial localization.",
	author = "Shervin Ardeshir, Amir Roshan Zamir, Alejandro Torroella, and Mubarak Shah.",
	journal = "In European Conference on Computer Vision–ECCV ",
	year = "2014", volume = "", number = "", pages = "602-617"}


@misc{CSS,
	author = {Farahat, Ahmed K., Ali Ghodsi, and Mohamed S. Kamel},
	title = {A fast greedy algorithm for generalized column subset selection.},
	note = {arXiv preprint arXiv:1312.6820 (2013).},
	year = 2014
}


@misc{DPM1,
 author = "Girshick, R. B. and Felzenszwalb, P. F. and McAllester, D.",
 title = "Discriminatively Trained Deformable Part Models, Release 5",
 howpublished = "http://people.cs.uchicago.edu/~rbg/latent-release5/"}

@article{DPM2,
  title = "Object Detection with Discriminatively Trained Part Based Models",
  author = "Felzenszwalb, P. F. and Girshick, R. B. and McAllester, D. and Ramanan, D.",
  journal = "IEEE Transactions on Pattern Analysis and Machine Intelligence",
  year = "2010", volume = "32", number = "9", pages = "1627--1645"}


@article{tensormatching,
  title={A tensor-based algorithm for high-order graph matching},
  author={Duchenne, Olivier and Bach, Francis and Kweon, In-So and Ponce, Jean},
  journal={Pattern Analysis and Machine Intelligence, IEEE Transactions on},
  volume={33},
  number={12},
  pages={2383--2395},
  year={2011},
  publisher={IEEE}
}

@inproceedings{uchiyama2009ar,
  title={AR GIS on a Physical Map Based on Map Image Retrieval Using LLAH Tracking.},
  author={Uchiyama, Hideaki and Saito, Hideo and Servieres, Myriam and Moreau, Guillaume and IRSTV, Ecole Centrale de Nantes-CERMA},
  booktitle={MVA},
  pages={382--385},
  year={2009}
}

@inproceedings{park2013tag,
  title={Tag configuration matcher for geo-tagging},
  author={Park, Minwoo and Chen, Yi and Shafique, Khurram},
  booktitle={Proceedings of the 21st ACM SIGSPATIAL International Conference on Advances in Geographic Information Systems},
  pages={374--377},
  year={2013},
  organization={ACM}
}

@incollection{li2012worldwide,
  title={Worldwide pose estimation using 3d point clouds},
  author={Li, Yunpeng and Snavely, Noah and Huttenlocher, Dan and Fua, Pascal},
  booktitle={Computer Vision--ECCV 2012},
  pages={15--29},
  year={2012},
  publisher={Springer}
}

@inproceedings{lee2013style,
  title={Style-aware mid-level representation for discovering visual connections in space and time},
  author={Lee, Yong Jae and Efros, Alexei A and Hebert, Martial},
  year={2013},
  organization={ICCV}
}

@INPROCEEDINGS{5206600, 
author={Lu Wang and Neumann, U.}, 
booktitle={Computer Vision and Pattern Recognition, 2009. CVPR 2009. IEEE Conference on}, 
title={A robust approach for automatic registration of aerial images with untextured aerial LiDAR data}, 
year={2009}, 
month={June}, 
pages={2623-2630}, 
keywords={image registration;image texture;optical radar;radar imaging;3D urban modeling;aerial images;airborne LiDAR technology;automatic registration;putative line segment;untextured aerial LiDAR data;Cameras;Cities and towns;Global Positioning System;Image generation;Image segmentation;Large-scale systems;Laser radar;Robustness;Urban planning;Videos}, 
doi={10.1109/CVPR.2009.5206600}, 
ISSN={1063-6919},}

@inproceedings{bioret2010towards,
  title={Towards outdoor localization from GIS data and 3D content extracted from videos},
  author={Bioret, Nicolas and Moreau, Guillaume and Servieres, Myriam},
  booktitle={Industrial Electronics (ISIE), 2010 IEEE International Symposium on},
  pages={3613--3618},
  year={2010},
  organization={IEEE}
}

@inproceedings{torralba2003context,
  title={Context-based vision system for place and object recognition},
  author={Torralba, Antonio and Murphy, Kevin P and Freeman, William T and Rubin, Mark A},
  booktitle={Computer Vision, 2003. Proceedings. Ninth IEEE International Conference on},
  pages={273--280},
  year={2003},
  organization={IEEE}
}

@article{torralba2003contextual,
  title={Contextual priming for object detection},
  author={Torralba, Antonio},
  journal={International Journal of Computer Vision},
  volume={53},
  number={2},
  pages={169--191},
  year={2003},
  publisher={Springer}
}

@ARTICLE{objectDetectionAssisted, 
author={Dasiopoulou, S. and Mezaris, V. and Kompatsiaris, I. and Papastathis, V.-K. and Strintzis, M.G.}, 
journal={Circuits and Systems for Video Technology, IEEE Transactions on}, 
title={Knowledge-assisted semantic video object detection}, 
year={2005}, 
month={Oct}, 
volume={15}, 
number={10}, 
pages={1210-1224}, 
keywords={multimedia systems;object detection;semantic Web;video signal processing;knowledge-assisted semantic video object detection;multimedia analysis;multimedia ontology;multimedia processing methods;semantic Web technologies;semantic annotation;transcoding systems;Bandwidth;Color;Computer vision;Context modeling;Independent component analysis;Knowledge representation;Object detection;Ontologies;Semantic Web;Transcoding;Knowledge-assisted analysis;multimedia ontologies;video analysis}, 
doi={10.1109/TCSVT.2005.854238}, 
ISSN={1051-8215},}


@INPROCEEDINGS{ShashuaGraphMatching, 
author={Zass, R. and Shashua, A.}, 
booktitle={Computer Vision and Pattern Recognition, 2008. CVPR 2008. IEEE Conference on}, 
title={Probabilistic graph and hypergraph matching}, 
year={2008}, 
month={June}, 
pages={1-8}, 
keywords={graph theory;image matching;image representation;image resolution;matrix algebra;probability;algebraic relation;convex optimization;hyper-edge representation;hyper-edge weight matrix;iterative successive projection algorithm;probabilistic graph-hypergraph matching;probabilistic setting;soft matching criterion;vertex-to-vertex probabilistic matching;Computer science;Image databases;Image recognition;Iterative algorithms;Object recognition;Optimal matching;Pixel;Projection algorithms;Scalability;Stochastic processes}, 
doi={10.1109/CVPR.2008.4587500}, 
ISSN={1063-6919},}

@INPROCEEDINGS{geo_reg_2, 
author={Kumar, Rakesh and Sawhney, H.S. and Asmuth, J.C. and Pope, A. and Hsu, S.}, 
booktitle={Pattern Recognition, 1998. Proceedings. Fourteenth International Conference on}, 
title={Registration of video to geo-referenced imagery}, 
year={1998}, 
month={Aug}, 
volume={2}, 
pages={1393-1400 vol.2}, 
keywords={image matching;image registration;image sequences;aerial mapping;aerial video imagery;annotations;camera coordinates;geo-coordinates;geo-referenced imagery;geo-spatial registration;monitoring systems;surveillance;target location;textual/graphical annotations;visually-guided navigation;Cameras;Image databases;Layout;Monitoring;Navigation;Surveillance;Target tracking;Terrain mapping;Visual databases;Visualization}, 
doi={10.1109/ICPR.1998.711963}, 
ISSN={1051-4651},}

@article{geo_reg_1,
  title={Image registration methods: a survey},
  author={Zitova, Barbara and Flusser, Jan},
  journal={Image and vision computing},
  volume={21},
  number={11},
  pages={977--1000},
  year={2003},
  publisher={Elsevier}
}

@misc{THUMOS13,
	author = "Jiang, Y. and Liu, J. and Roshan Zamir, A. and Laptev,
	I. and Piccardi, M. and Shah, M. and Sukthankar, R.",
	title = "{THUMOS} Challenge: Action Recognition with a Large
	Number of Classes",
	howpublished = "\url{http://crcv.ucf.edu/ICCV13-Action-Workshop/}",
	Year = {2013}}

	
	
@inproceedings{UCF101,
	author = {Soomro, K. and Roshan Zamir, A. and Shah, M.},
	booktitle = {CRCV-TR-12-01},
	title = {{UCF101}: A Dataset of 101 Human Actions Classes From
	Videos in The Wild},
	year = {2012}}

@incollection{ZamirECCV10,
  title={Accurate image localization based on google maps street view},
  author={Zamir, Amir Roshan and Shah, Mubarak},
  booktitle={European Conference on Computer Vision (ECCV)},
  pages={255--268},
  year={2010},
  publisher={Springer}
}

@ARTICLE{ZamirPAMI13, 
author={Amir Roshan Zamir and Mubarak Shah}, 
journal={IEEE Transactions on Pattern Analysis and Machine Intelligence (T-PAMI)}, 
title={Image Geo-localization Based on Multiple Nearest Neighbor Feature Matching Using Generalized Graphs}, 
year={2014}, 
}




@inproceedings{ZamirECCV12,
   author = "A. Roshan Zamir and A. Dehghan and M. Shah",
   title = "{GMCP-Tracker}: Global Multi-object Tracking Using Generalized Minimum Clique Graphs",
   booktitle = "European Conference on Computer Vision (ECCV)",
   year = "2012",
}

@inproceedings{TagrefinementCVPR14,
   Author = {Amir Roshan Zamir and Shervin Ardeshir and Mubarak Shah},
   Booktitle = {IEEE International Conference on Computer Vision and Pattern Recognition (CVPR)},
   Title = {Robust Refinement of GPS-Tags Using Random Walks with an Adaptive Damping Factor},
   Year = {2014}}

@inproceedings{GMCPVideo14,
   Author = {Shayan Modiri and Amir Roshan Zamir and Mubarak Shah},
   Booktitle = {IEEE International Conference on Computer Vision and Pattern Recognition (CVPR)},
   Title = {Video Classification Based on Generalized Maximum Co-occurrence Cliques},
   Year = {2014}}

@inproceedings{VacaZamir12,
   Author = {Gonzalo Vaca and Amir Roshan Zamir and Mubarak Shah},
   Booktitle = {IEEE International Conference on Computer Vision and Pattern Recognition (CVPR)},
   Title = {City Scale Geo-spatial Trajectory Estimation of a Moving Camera},
   Year = {2012}}


@inproceedings{ZamirACMMM13,
   Author = {Amir Roshan Zamir and Afshin Dehghan and Mubarak Shah},
   Booktitle = {Proceeding of ACM International Conference on Multimedia ({ACM MM})},
   Title = {Visual Business Recognition - A Multimodal Approach},
   Year = {2013}}
   
 
   

@inproceedings{ZamirICMLA2011,
  title={Street view challenge: Identification of commercial entities in street view imagery},
  author={Zamir, Amir Roshan and Darino, Alexander and Shah, Mubarak},
  booktitle={2011 10th International Conference on Machine Learning and Applications (ICMLA)},
  volume={2},
  pages={380--383},
  year={2011},
  organization={IEEE}
}


%%%%%%%%%%PAMI13




@book{clusteranalysis,
 author = {Everitt, Brian S. and Landau, Sabine and Leese, Morven},
 title = {Cluster Analysis},
 year = {2009},
 isbn = {0340761199, 9780340761199},
 edition = {4th},
 publisher = {Wiley Publishing},
}

@inproceedings{huberloss,
   author = "PJ Huber",
   title = "Robust estimation of a location parameter",
   booktitle = "The Annals of Mathematical Statistics",
   year = "1964",
}


@inproceedings{flickr,
   author = "",
   title = "Flickr - Photo Sharing: http://www.flickr.com",
   booktitle = "",
   year = "",
}

@inproceedings{panoramio,
   author = "",
   title = "Panoramio - Photos of the World: http://www.panoramio.com",
   booktitle = "",
   year = "",
}

@inproceedings{Picasa,
   author = "",
   title = "picasa: picasa.google.com",
   booktitle = "",
   year = "",
}

@inproceedings{videogoogle,
   author = "J. Sivic and A. Zisserman",
   title = "Video Google: a text retrieval approach to object matching in videos",
   booktitle = "International Conference on Computer Vision",
   year = "2003",
}

@inproceedings{IM2GPS,
   author = "J. Hays and A. Efros",
   title = "im2gps: estimating geographic information from a single image",
   booktitle = "IEEE Conference on Computer Vision and Pattern Recognition",
   year = "2008",
}


@inproceedings{cityscale,
   author = "G. Schindler and M. Brown and and R. Szeliski",
   title = "City-Scale Location Recognition",
   booktitle = "IEEE Conference on Computer Vision and Pattern Recognition",
   year = "2007",
}

@inproceedings{GSVSivic,
   author = "J. Knopp and J. Sivic and T.Pajdla",
   title = "Avoding confusing features in place recognition",
   booktitle = "European Conference on Computer Vision",
   year = "2010",
}

@inproceedings{torsten11,
   author = "T. Sattler and B. Leibe and and L. Kobbelt",
   title = "Fast image-based localization using direct 2d-to-3d matching",
   booktitle = "International Conference on Computer Vision",
   year = "2010",
}

@inproceedings{2D3Dprior,
   author = "Y. Li and N. Snavely and D. Huttenlocher",
   title = "Location Recognition Using Prioritized Feature Matching",
   booktitle = "European Conference on Computer Vision",
   year = "2010",
}

@inproceedings{torsten12,
   author = "T. Sattler and B. Leibe and and L. Kobbelt",
   title = "Improving Image-Based Localization by Active Correspondence Search",
   booktitle = "European Conference on Computer Vision",
   year = "2012",
}


@inproceedings{zheng2009tour,
  title={Tour the world: building a web-scale landmark recognition engine},
  author={Zheng, Yan-Tao and Zhao, Ming and Song, Yang and Adam, Hartwig and Buddemeier, Ulrich and Bissacco, Alessandro and Brucher, Fernando and Chua, T-S and Neven, Hartmut},
   booktitle = "IEEE Conference on Computer Vision and Pattern Recognition",
  pages={1085--1092},
  year={2009},
  organization={IEEE}
}

@inproceedings{3Dphrase,
   author = "Q. Hao and R. Cai and Z. Li and L. Zhang and Y. Pang and and F. Wu",
   title = "3D Visual Phrases for Landmark Recognition",
   booktitle = "IEEE Conference on Computer Vision and Pattern Recognition",
   year = "2012",
}

@inproceedings{visualphrase,
   author = "Y. Zhang and Z Jia and T. Chen",
   title = "Image retrieval with geometry-preserving visual phrases",
   booktitle = "IEEE Conference on Computer Vision and Pattern Recognition",
   year = "2011",
}

@inproceedings{globalsift,
   author = "E.N. Mortensen and Hongli Deng and L. Shapiro",
   title = "A SIFT descriptor with global context",
   booktitle = "IEEE Conference on Computer Vision and Pattern Recognition",
   year = "2005",
}

@inproceedings{shapesift,
   author = "K. Mikolajczyk and A. Zisserman and C. Schmid",
   title = "Shape recognition with edge-based features",
   booktitle = "British Machine Vision Conference",
   year = "2003",
}


@inproceedings{asiftglobal,
   author = "B. Cao and C. Ma and Z. Liu",
   title = "Affine-invariant SIFT descriptor with global context",
   booktitle = "International Congress on Image and Signal Processing",
   year = "2010",
}

@inproceedings{lostinquantization,
   author = "J. Philbin and O. Chum and and M. Isard and J. Sivic and and A. Zisserman",
   title = "Lost in Quantization: Improving Particular Object Retrieval in Large Scale Image Databases",
   booktitle = "IEEE Conference on Computer Vision and Pattern Recognition",
   year = "2008",
}



@inproceedings{FAP,
   author = "A. Koster and  S. Van Hoesel and  A. Kolen",
   title = "The partial constraint satisfaction problem: Facets and lifting theorems",
   booktitle = "Operations Research Letters 23",
   year = "1998",
}



@inproceedings{linkagecluster,
   author = "J. C. Gower and G. J. S. Ross",
   title = "Minimum Spanning Trees and Single Linkage Cluster Analysis",
   booktitle = "Journal of the Royal Statistical Society",
   year = "1969",
}


@inproceedings{GMST-Tabusearch,
   author = "Z. Wang and C.H and Chev A. Lim",
   title = "Tabu search for generalized minimum spanning tree problem",
   booktitle = "Pacific Rim International Conference on Artificial Intelligence",
   year = "2006",
}



@inproceedings{KDTREE,
   author = "J. Bentley",
   title = "Multidimensional binary search trees in database applications",
   booktitle = "IEEE Transactions on Software Engineering",
   year = "1979",
}



@inproceedings{GNDP,
   author = "C. Feremans and M. Labbe and G. Laporte",
   title = "Generalized network design problems",
   booktitle = "European Journal of Operational Research Volume 148, Issue 1",
   year = "2003",
}


@inproceedings{SIFT,
   author = "D. G. Lowe",
   title = "Distinctive image features from scale-invariant keypoints",
   booktitle = "International Journal of Computer Vision",
   year = "2004",
}

@inproceedings{FLANN,
   author = "M. Muja and D. G. Lowe",
   title = "Fast approximate nearest neighbors with automatic algorithm",
   booktitle = "International Conference on Computer Vision Theory and Applications",
   year = "2009",
}



@inproceedings{myunggmst,
   author = "Y. S. Myung and C. H. Lee and D. W. Tcha",
   title = "On the generalized minimum spanning tree problem",
   booktitle = "Networks",
   year = "1995",
}



@inproceedings{GIST,
   author = "A. Torralba",
   title = "Contextual priming for object detection",
   booktitle = "International Journal of Computer Vision, Vol. 53(2), 169-191",
   year = "2003",
}


@inproceedings{globalgeomtry1,
   author = "Y. Avrithis and G. Tolias and Y. Kalantidis",
   title = "Feature map hashing: sub-linear indexing of appearance and global geometry",
   booktitle = "International Conference on Multimedia",
   year = "2010",
}



@inproceedings{philbin,
   author = "J. Philbin and O. Chum and M. Isard and J. Sivic and A. Zisserman",
   title = "Object retrieval with large vocabularies and fast spatial matching",
   booktitle = "IEEE Conference on Computer Vision and Pattern Recognition",
   year = "2007",
}



@inproceedings{kerneldistance,
   author = "G. Wu and E. Y. Chang and N. Panda",
   title = "Formulating Distance Functions via the Kernel Trick",
   booktitle = "In Conf. on Knowledge Discovery and Data Mining",
   year = "2005",
}




@inproceedings{GMSTTABU,
   author = "D. Ghosh",
   title = "Solving medium to large sized Euclidean generalized minimum spanning tree problems",
   booktitle = "WP. No. 2003-08-02, Indian Institute of Management",
   year = "2003",
}


@inproceedings{hakeem,
   author = "A. Hakeem and R. Vezzani and M. Shah and R. Cucchaira",
   title = "Estimating Geospatial Trajectory of a Moving Camera",
   booktitle = "International Conference on Pattern Recognition",
   year = "2006",
}


@inproceedings{protein,
   author = "E. Althaus and O. Kohlbacher and H. Lenhof and P. M�ller.",
   title = "A combinatorial approach to protein docking with flexible side chains",
   booktitle = "Journal of Computational Biology",
   year = "2002",
}


@inproceedings{LoopyBP,
   author = "J. Pearl",
   title = "Probabilistic Reasoning in Intelligent Systems: Networks of Plausble Inference",
   booktitle = "Intelligent Systems: Networks of Plausble Inference",
   year = "1988",
}




@inproceedings{SCRAMSAC,
   author = "T. Sattler and B. Leibe and L. Kobbelt",
   title = "SCRAMSAC: Improving RANSAC`s Efficiency with a Spatial Consistency Filter",
   booktitle = "International Conference on Computer Vision",
   year = "2009",
}


@inproceedings{LoopyBPloos,
   author = "Y. Weiss",
   title = "Correctness of Local Probability Propagation in Graphical Models with Loop",
   booktitle = "Neural Computation",
   year = "2000",
}



@inproceedings{san_fran_dataset,
   author = "D. Chen and G. Baatz and K. Koeser and S. Tsai and R. Vedantham and T. Pylvanainen and K. Roimela and X. Chen and J. Bach and M. Pollefeys and B. Girod and R. Grzeszczuk",
   title = "City-scale landmark identification on mobile devices",
   booktitle = "IEEE International Conference on Computer Vision and Pattern Recognition",
   year = "2011",

}


@inproceedings{MLSAC,
   author = "P.H.S. Torr and A. Zisserman",
   title = "A New Robust Estimator with Application to Estimating Image Geometry",
   booktitle = "IEEE Computer Vision and Image Understanding",
   year = "2000",
}


@inproceedings{MSAC,
   author = "H. Wang and D. Mirota and G.D. Hager",
   title = "A Generalized Kernel Consensus-Based Robust Estimator",
   booktitle = "IEEE Transactions on Pattern Analysis and Machine Intelligence",
   year = "2010",
}


@inproceedings{akihinoCVPR13,
   author = "A. Torii and J. Sivic and T. Pajdla and M. Okutomi",
   title = "Visual Place Recognition with Repetitive Structures",
   booktitle = "IEEE International Conference on Computer Vision and Pattern Recognition",
   year = "2013",
}




@inproceedings{JegouCVPR09,
   author = "E. Jegou and M. Douze and C. Schmid",
   title = "On the burstiness of visual elements",
   booktitle = "IEEE International Conference on Computer Vision and Pattern Recognition",
   year = "2009",
}

%%%%%%%%%%%%%%%%%%%%%%%%%%%ECCV10


 

@inproceedings{urban,
 author = {Wei Zhang and Jana Kosecka},
 title = {Image Based Localization in Urban Environments},
 booktitle = {3DPVT '06: Proceedings of the Third International Symposium on 3D Data Processing, Visualization, and Transmission (3DPVT'06)},
 pages = {33--40},
 year = {2006}
 }

@inproceedings{rome,
  author = {S. Agarwal and N. Snavely and I.  Simon and S. M. Seitz and R. Szeliski},
  title = {Building Rome in a Day},
  booktitle = iccv,
  year = 2009
}
@inproceedings{statcam,
  author={Nathan Jacobs and Scott Satkin and Nathaniel Roman and Richard Speyer and Robert Pless},
  title={Geolocating Static Cameras},
  booktitle=iccv,
  year=2007
}
% Features
@article{SIFT,
  author = {D. G. Lowe},
  title = {Distinctive Image Features from scale-invariant keypoints},
  journal = ijcv,
  volume = 60,
  number = 2,
  pages2 = {91--110},
  year = 2004
}
@inproceedings{human,
  author = {E. Kalogerakis and O. Vesselova and J. Hays and A. Efros and A. Hertzmann},
  booktitle = iccv,
  title = {Image Sequence Geolocation with Human Travel Priors},
  year = {2009}
}
@inproceedings{world,
    author = {D. Crandall and L. Backstrom and D. Huttenlocher and J. Kleinberg},
    booktitle = {International World Wide Web Conference},
    title = {Mapping the World's Photos},
    year = {2009}
}

@misc{harris,
    author = {Konstantinos G. Derpanis},
    title = {The Harris Corner Detector},
    year = {2004}
}
@article{COLSIFT,
  author = {K. van de Sande and others},
  author2 = {K. van de Sande and T. Gevers and C. Snoek},
  title  = {Evaluating Color Descriptors for Object and Scene Recognition},
  journal= pami,
  year   = 2010
}
@inproceedings{FLANN,
  author = {Marius Muja and David G. Lowe},
  title = {Fast Approximate Nearest Neighbors with Automatic Algorithm Configuration},
  booktitle = {VISAPP},
  year = 2009
}

    @article{kurtosis,
     title = {Kurtosis: A Critical Review},
     author = {Balanda, Kevin P. and MacGillivray, H. L.},
     journal = {The American Statistician},
     volume = {42},
     number = {2},
     pages = {111--119},
     year = {1988},
    }
    
@ARTICLE{kurtosisapp,
  AUTHOR =       {R.B. Randall},
  TITLE =        {Applications of Spectral Kurtosis in Machine Diagnostics and Prognostics},
  JOURNAL =      { 	Key Engineering Materials },
  YEAR =         {2005},
  volume =       { 293-294},
  pages =        {21-32},

}
@INPROCEEDINGS{generalizedkurtosis,
    author = {Andre Tkacenko and P. P. Vaidyanathan},
    title = {Generalized Kurtosis and Applications in Blind Equalization of MIMO Channels},
    booktitle = {Proc. IEEE Asilomar Conference on Signals, Systems, and Computers},
    year = {2001}
}





@article{hakeem,
author = {Asaad Hakeem and Roberto Vezzani and Mubarak Shah and Rita Cucchiara},
title = {Estimating Geospatial Trajectory of a Moving Camera},
journal ={Pattern Recognition, International Conference on},volume = {2},issn = {1051-4651},year = {2006},pages = {82-87},doi = {http://doi.ieeecomputersociety.org/10.1109/ICPR.2006.499},publisher = {IEEE Computer Society},address = {Los Alamitos, CA, USA},
}


@inproceedings{Nister,
 author = {Nister, David and Stewenius, Henrik},
 title = {Scalable Recognition with a Vocabulary Tree},
 booktitle = {Proceedings of the 2006 IEEE Computer Society Conference on Computer Vision and Pattern Recognition - Volume 2},
 series = {CVPR '06},
 year = {2006},
 isbn = {0-7695-2597-0},
 pages = {2161--2168},
 numpages = {8},
 url = {http://dx.doi.org/10.1109/CVPR.2006.264},
 doi = {http://dx.doi.org/10.1109/CVPR.2006.264},
 acmid = {1153548},
 publisher = {IEEE Computer Society},
 address = {Washington, DC, USA},
}

@article{Photosynth,
 author = {Snavely, Noah and Seitz, Steven M. and Szeliski, Richard},
 title = {Photo tourism: exploring photo collections in 3D},
 journal = {ACM Trans. Graph.},
 volume = {25},
 issue = {3},
 month = {July},
 year = {2006},
 issn = {0730-0301},
 pages = {835--846},
 numpages = {12},
 url = {http://doi.acm.org/10.1145/1141911.1141964},
 doi = {http://doi.acm.org/10.1145/1141911.1141964},
 acmid = {1141964},
 publisher = {ACM},
 address = {New York, NY, USA},
 keywords = {image-based modeling, image-based rendering, photo browsing, structure from motion},
}


@misc{vlfeat,
 Author = {A. Vedaldi and B. Fulkerson},
 Title = {{VLFeat}: An Open and Portable Library 
          of Computer Vision Algorithms},
 Year  = {2008},
 Howpublished = {\url{http://www.vlfeat.org/}}
}




%%%%%%%%%%CVPR12


@ARTICLE{Civera2010,
  author = {J. Civera and O. Grasa and A. Davison and J. Montiel},
  title = {1-point ransac for ekf filtering: Application to real-time structure
	from motion and visual odometry},
  journal = {Journal of Field Robotics},
  year = {2010},
  volume = {27},
  pages = {609-631},
  owner = {Student},
  timestamp = {2011.11.18}
}

@ARTICLE{Cummins2010,
  author = {Mark Cummins and Paul Newman},
  title = {Appearance-only SLAM at large scale with FAB-MAP 2.0},
  journal = {International Journal of Robotics Research},
  year = {2010},
  month = {November},
  owner = {Student},
  timestamp = {2012.03.18}
}

@ARTICLE{Cummins2008,
  author = {Mark Cummins and Paul Newman},
  title = {FAB-MAP: Probabilistic Localization and Mapping in the Space of Appearance},
  journal = {International Journal of Robotics Research},
  year = {2008},
  month = {June},
  owner = {Student},
  timestamp = {2012.03.18}
}

@CONFERENCE{Davison2003,
  author = {A. Davison},
  title = {Real-time simultaneous localisation and mapping with a single camera},
  booktitle = {ICCV},
  year = {2003},
  owner = {Student},
  timestamp = {2011.11.18}
}

@CONFERENCE{vehicle,
  author = {F. Diego and D. Ponsa and J. Serrat and A.M. Lo?pez},
  title = {Vehicle Geolocalization based on video synchronization},
  booktitle = {Intelligent Transportation Systems (ITSC)},
  year = {2010},
  owner = {Student},
  timestamp = {2011.11.16}
}

@ARTICLE{Durrant-Whyte2006,
  author = {H. Durrant-Whyte and T. Bailey},
  title = {Simultaneous localization and mapping (slam): Part i the essential
	algorithms},
  journal = {Robotics and Automation Magazine},
  year = {2006},
  volume = {13},
  pages = {99-110},
  number = {2},
  owner = {Student},
  timestamp = {2011.11.18}
}

@ARTICLE{MSTmorph,
  author = {Luiz Henrique de Figueiredo and Jonas de Miranda Gomes},
  title = {Computational morphology of curves},
  journal = {The Visual Computer},
  year = {1994},
  volume = {11},
  pages = {105-112},
  number = {2},
  owner = {Student},
  timestamp = {2011.11.16}
}

@UNPUBLISHED{Google,
  author = {Google},
  title = {http://maps.google.com},
  owner = {Student},
  timestamp = {2011.11.18}
}

@UNPUBLISHED{youtube,
  author = {Google},
  title = {http://www.youtube.com},
  owner = {Student},
  timestamp = {2011.11.18}
}

@CONFERENCE{Hakeem2006,
  author = {A. Hakeem and R. Vezzani and M. Shah and R. Cucchiara},
  title = {Estimating Geospatial Trajectory of a Moving Camera},
  booktitle = {ICPR},
  year = {2006},
  journal = {18th International Conference on Pattern Recognition (ICPR)},
  owner = {Student},
  timestamp = {2011.11.16}
}

@CONFERENCE{realtimester,
  author = {A. Howard},
  title = {Real-time stereo visual odometry for autonomous ground vehicles},
  booktitle = {IROS},
  year = {2008},
  owner = {Student},
  timestamp = {2011.11.16}
}

@UNPUBLISHED{FlickerMap,
  author = {http://www.flickr.com},
  title = {Flickr},
  howpublished = {http://www.flickr.com/},
  owner = {Student},
  timestamp = {2011.11.18},
  url = {http://www.flickr.com/}
}

@CONFERENCE{Isard1998,
  author = {M. Isard and A. Blake},
  title = {A smoothing filter for Condensation},
  booktitle = {European Conference in Computer Vision (ECCV)},
  year = {1998},
  owner = {Student},
  timestamp = {2011.11.16}
}

@ARTICLE{Kitawa,
  author = {G Kitawa},
  title = {Monte Carlo Filter and Smoother for Non-Gaussian Nonlinear State
	space Models},
  journal = {Journal of Computational and Graphical Statistics},
  year = {1996},
  volume = {5},
  pages = {1-25},
  number = {1},
  month = {March},
  owner = {Student},
  timestamp = {2011.11.16}
}

@CONFERENCE{Klein2007,
  author = {G. Klein and D. Murray},
  title = {Parallel tracking and mapping for smaller workspaces},
  booktitle = {International Symposium on Mixed and Augmented Reality},
  year = {2007},
  owner = {Student},
  timestamp = {2011.11.18}
}

@ARTICLE{curverecMST,
  author = {In-Kwon Lee},
  title = {Curve reconstruction from unorganized points},
  journal = {Computer Aided Geometric Design},
  year = {2000},
  volume = {17},
  pages = {161-177},
  owner = {Student},
  timestamp = {2011.11.16}
}

@CONFERENCE{2D3DICCV11,
  author = {Yangyan Li and Qian Zheng and Andrei Sharf and Daniel Cohen-Or and
	Baoquan Chen and Niloy J. Mitra},
  title = {2D-3D Fusion for Layer Decomposition of Urban Facades},
  booktitle = {International Conference in Computer Vision (ICCV)},
  year = {2011},
  owner = {Student},
  timestamp = {2011.11.18}
}

@ARTICLE{Lowe2004,
  author = {David G. Lowe},
  title = {Distinctive image features from scale-invariant keypoints},
  journal = {IJCV},
  year = {2004},
  number = {2},
  owner = {Student},
  timestamp = {2011.11.16}
}

@CONFERENCE{MSTReg,
  author = {B. Ma and A. Hero and J. Gorman and O. Michel},
  title = {Image registration with minimum spanning tree algorithm},
  booktitle = {International Conference on Image Processing},
  year = {2000},
  owner = {Student},
  timestamp = {2011.11.16}
}

@CONFERENCE{realtimeloc,
  author = {E. Mouragnon and M. Lhuillier and M. Dhome and F. Dekeyser and P.
	Sayd},
  title = {Real time localization and 3d reconstruction},
  booktitle = {Computer Vision and Pattern Recognition (CVPR)},
  year = {2006},
  owner = {Student},
  timestamp = {2011.11.16}
}

@CONFERENCE{Muja2009,
  author = {M. Muja and D. Lowe},
  title = {Fast approximate nearest neighbors with automatic algorithm configuration.},
  booktitle = {VISAPP'09},
  owner = {Student},
  timestamp = {2011.11.16}
}

@CONFERENCE{Nister2004,
  author = {D. Nister and O. Naroditsky and J. Bergen},
  title = {Visual odometry},
  booktitle = {CVPR},
  year = {2004},
  owner = {Student},
  timestamp = {2011.11.18}
}



@CONFERENCE{MSTNetwork,
  author = {Radia Perlman},
  title = {An algorithm for distributed computation of a spanningtree in an
	extended LAN},
  booktitle = {SIGCOMM},
  year = {1985},
  owner = {Student},
  timestamp = {2011.11.16}
}

@CONFERENCE{TorstenSattler2011,
  author = {Torsten Sattler and Bastian Leibe and Leif Kobbelt},
  title = {Fast Image-Based Localization using Direct 2D-to-3D Matching},
  booktitle = {ICCV'11},
  owner = {Student},
  timestamp = {2012.04.05}
}

@CONFERENCE{fast,
  author = {Torsten Sattler and Bastian Leibe and Leif Kobbelt},
  title = {Fast Image-Based Localization using Direct 2D-to-3D Matching},
  booktitle = {IEEE International Conference on Computer Vision (ICCV 11)},
  year = {2011},
  owner = {Student},
  timestamp = {2011.11.16}
}

@ARTICLE{Scaramuzza2010,
  author = {Davide Scaramuzza},
  title = {1-Point-RANSAC Structure from Motion for Vehicle-Mounted Cameras
	by Exploiting Non-holonomic Constraints},
  journal = {IJCV},
  year = {2010},
  volume = {95},
  number = {1},
  owner = {Student},
  timestamp = {2011.11.21}
}

@ARTICLE{Scaramuzza2011,
  author = {D. Scaramuzza and F. Fraundorfer},
  title = {Visual Odometry: Part I - The First 30 Years and Fundamentals},
  journal = {IEEE Robotics and Automation Magazine},
  year = {2011},
  volume = {18},
  number = {4},
  month = {December},
  owner = {Student},
  timestamp = {2011.11.18}
}



@CONFERENCE{Tardif2008,
  author = {J. Tardif and Y. Pavlidis and K. Daniilidis},
  title = {Monocular visual odometry in urban environments using an omnidirectional
	camera},
  booktitle = {International Conference on Intelligent Robots and Systems},
  year = {2008},
  owner = {Student},
  timestamp = {2011.11.18}
}

@CONFERENCE{Torralba2003,
  author = {A. Torralba and K. P. Murphy and W. T. Freeman and M. A. Rubin},
  title = {Context-based vision system for place and object recognition},
  booktitle = {ICCV},
  year = {2003},
  owner = {Student},
  timestamp = {2012.03.18}
}







%%%%%%ACM MM13


@inproceedings{GoogleGoggle,
   author = "",
   title = "Google Goggles, \emph{www.google.com/mobile/goggles/}",
   booktitle = "",
   year = "",
}

 

@inproceedings{nokia,
   author = "",
   title = "Nokia City Lens, \emph{http://betalabs.nokia.com/trials/nokia-city-lens-for-windows-phone}",
   booktitle = "",
   year = "",
}

@inproceedings{GNUOCR,
   author = "",
   title = "GNU-Orcad OCR, \emph{http://www.gnu.org/s/ocrad/}",
   booktitle = "",
   year = "",
}


@inproceedings{YellowPages,
   author = "",
   title = "YellowPages, \emph{http://www.yellowpages.com/}",
   booktitle = "",
   year = "",
}

@inproceedings{Yelp,
   author = "",
   title = "Yelp, \emph{http://www.yelp.com/}",
   booktitle = "",
   year = "",
}

@inproceedings{WPS,
   author = "",
   title = "Wi-Fi positioning system, \emph{http://www.skyhookwireless.com/howitworks/}",
   booktitle = "",
   year = "",
}
 
 @inproceedings{NaturalSceneText,
    author = "X. Chen and J. Yang and J. Zhang and A. Waibel",
    title = "Automatic Detection and Recognition of Signs From Natural Scenes",
    booktitle = "IEEE TRANSACTIONS ON IMAGE PROCESSING",
    year = "2004",
}

 
 @inproceedings{UCLANaturalScene,
    author = "X. Chen and A. Yuille",
    title = "Detecting and Reading Text in
Natural Scenes",
    booktitle = "IEEE Conference on Computer Vision and Pattern Recognition",
    year = "2004",
}
 

 @inproceedings{RoadSign,
    author = "X. Chen; J. Yang; J. Zhang; A. Waibel",
    title = "Automatic detection and recognition of signs from natural scenes",
    booktitle = "IEEE Transactions on Image Processing",
    year = "2004",
}
 
 @inproceedings{3DStreetView,
    author = "B. Micusik; J. Kosecka",
    title = "Piecewise planar city 3D modeling from street view panoramic sequences",
    booktitle = "IEEE Conference on Computer Vision and Pattern Recognition",
    year = "2009",
}
 

@inproceedings{3DStreetView,
    author = "J. Knopp and J. Sivic and T. Pajdla",
    title = "Avoiding confusing features in place recognition",
    booktitle = "European Conference on Computer Vision",
    year = "2010",
}

 
@inproceedings{3DStreetView,
    author = "J. Knopp and J. Sivic and T. Pajdla",
    title = "Avoiding confusing features in place recognition",
    booktitle = "European Conference on Computer Vision",
    year = "2010",
}

@ARTICLE{STRB, 
author={Weinman, J.J. and Learned-Miller, E. and Hanson, A.R.}, 
journal={Pattern Analysis and Machine Intelligence, IEEE Transactions on}, 
title={Scene Text Recognition Using Similarity and a Lexicon with Sparse Belief Propagation}, 
year={2009}, 
volume={31}, 
number={10}, 
pages={1733-1746}, 
keywords={Markov processes;character recognition;image recognition;STR;character image;document recognition;lexicon;scene text recognition;sparse belief propagation;Scene text recognition;belief propagation;conditional random fields;factor graphs;graphical models;language model;lexicon;optical character recognition;similarity;sparse belief propagation.}, 
doi={10.1109/TPAMI.2009.38}, 
ISSN={0162-8828},}
 

 @INPROCEEDINGS{ICCV11, 
 author={Kai Wang and Babenko, B. and Belongie, S.}, 
 booktitle={Computer Vision (ICCV), 2011 IEEE International Conference on}, 
 title={End-to-end scene text recognition}, 
 year={2011}, 
 pages={1457-1464}, 
 keywords={computer vision;object recognition;optical character recognition;text detection;OCR engine;computer vision community;domain-specific methods;end-to-end scene text recognition;image word recognition;natural images;object recognition;text detection;word detection;Detectors;Image recognition;Object recognition;Optical character recognition software;Pipelines;Text recognition;Training}, 
 doi={10.1109/ICCV.2011.6126402}, 
ISSN={1550-5499},}
 

@INPROCEEDINGS{SWT, 
author={Epshtein, B. and Ofek, E. and Wexler, Y.}, 
booktitle={Computer Vision and Pattern Recognition (CVPR), 2010 IEEE Conference on}, 
title={Detecting text in natural scenes with stroke width transform}, 
year={2010}, 
pages={2963-2970}, 
keywords={image processing;text analysis;image operator;image pixel;natural images;natural scenes;stroke width transform;text detection;Colored noise;Computer vision;Engines;Filter bank;Geometry;Image segmentation;Layout;Optical character recognition software;Pixel;Robustness}, 
doi={10.1109/CVPR.2010.5540041}, 
ISSN={1063-6919},}

 @ARTICLE{apptemplate, 
 author={Jain, A.K. and Yu Zhong and Lakshmanan, S.}, 
 journal={Pattern Analysis and Machine Intelligence, IEEE Transactions on}, 
 title={Object matching using deformable templates}, 
 year={1996}, 
 volume={18}, 
 number={3}, 
 pages={267-278}, 
 keywords={Bayes methods;image matching;image segmentation;object recognition;probability;visual databases;Bayesian scheme;deformable templates;edge information;image database;image segmentation;object localization;object matching;object shape;optimisation;probabilistic deformation;Bayesian methods;Computational efficiency;Computer science;Image databases;Image retrieval;Image segmentation;Information retrieval;Object recognition;Prototypes;Shape}, 
 doi={10.1109/34.485555}, 
ISSN={0162-8828},}


@article{OCRtemplate,
  title={Template-based online character recognition},
  author={Connell, Scott D and Jain, Anil K},
  journal={Pattern Recognition},
  volume={34},
  number={1},
  pages={1--14},
  year={2001},
  publisher={Elsevier}
}


@inproceedings{ABBYY,
    author = "",
    title = "http://www.abbyy.com/",
    booktitle = "",
    year = "",
}

 @inproceedings{MobileL1,
   title={Mobile product recognition},
   author={Tsai, Sam S and Chen, David and Chandrasekhar, Vijay and Takacs, Gabriel and Cheung, Ngai-Man and Vedantham, Ramakrishna and Grzeszczuk, Radek and Girod, Bernd},
   booktitle={Proceedings of the international conference on Multimedia},
   pages={1587--1590},
   year={2010},
   organization={ACM}
}

@inproceedings{MobileL2,
  title={Building book inventories using smartphones},
  author={Chen, David M and Tsai, Sam S and Girod, Bernd and Hsu, Cheng-Hsin and Kim, Kyu-Han and Singh, Jatinder Pal},
  booktitle={Proceedings of the international conference on Multimedia},
  pages={651--654},
  year={2010},
  organization={ACM}
}



@book{huber2011robust,
  title={Robust statistics},
  author={Huber, Peter J},
  year={2011},
  publisher={Springer}
}



@inproceedings{visualSFM1,
   author = "C. Wu and S. Agarwal and B. Curless and S. M. Seitz",
   title = "Multicore Bundle Adjustment",
   booktitle = "IEEE International Conference on Computer Vision and Pattern Recognition",
   year = "2011",
}



@article{visualSFM2,
  title={VisualSFM: A Visual Structure from Motion System},
  author={Ch. Wu},
  journal={http://ccwu.me/vsfm/},
  year = "2011",

}



@book{RWbook,
  title={The Theory of Stochastic Processes },
  author={Frank Spitzer},
  year={2001},
  publisher={Springer}
}


@article{visualrankRW,
  title={Visualrank: Applying pagerank to large-scale image search},
  author={Yushi Jing and Shumeet Baluja},
   booktitle = "IEEE Transactions on Pattern Analysis and Machine Intelligence",
  year={2008},
}

@inproceedings{aerial_localization_harpeet,
  title={Geo-localization of street views with aerial image databases},
  author={Bansal, Mayank and Sawhney, Harpreet S and Cheng, Hui and Daniilidis, Kostas},
  booktitle={Proceedings of the 19th ACM international conference on Multimedia},
  pages={1125--1128},
  year={2011},
  organization={ACM}
}


@inproceedings{linCVPR13,
  author={Tsung-Yi  Lin and Serge Belongie and James Hays},
  title={Cross-View Image Geolocalization},
   booktitle = "IEEE International Conference on Computer Vision and Pattern Recognition",
   year = "2013",

}



@inproceedings{liu2010image,
  author={Dong Liu and Xian-Sheng Hua and Meng Wang and Hong-Jiang Zhang},
  title={Image retagging},
   booktitle = "International Conference on Multimedia",
   year = "2010",

}


@inproceedings{tagrankingRW,
  author={Dong  Liu and Xian-Sheng Hua and Linjun Yang and Meng Wang and Hong-Jiang Zhang},
  title={Tag ranking},
   booktitle = "International Conference on Multimedia",
   year = "2009",

}


@inproceedings{zhu2010image,
  author={Guangyu Zhu and Shuicheng Yan and Yi Ma },
  title={Image tag refinement towards low-rank, content-tag prior and error sparsity},
   booktitle = "International Conference on Multimedia",
   year = "2010",

}

@inproceedings{TagRelevance, 
author={Xirong Li and G M Snoekand and Marcel Worring}, 
title={Learning Social Tag Relevance by Neighbor Voting}, 
booktitle={IEEE Transactions on Multimedia}, 
year={2009}, 
}




@inproceedings{toriiCVPR13,
  author={A. Torii and J. Sivic and T. Pajdla and M. Okutomi},
  title={Visual place recognition with repetitive structures},
   booktitle = "IEEE International Conference on Computer Vision and Pattern Recognition",
   year = "2013",

}


@inproceedings{snavely2011discrete,
  title={Discrete-continuous optimization for large-scale structure from motion},
  author={Crandall, David and Owens, Andrew and Snavely, Noah and Huttenlocher, Dan},
  booktitle={Computer Vision and Pattern Recognition (CVPR), 2011 IEEE Conference on},
  pages={3001--3008},
  year={2011},
}


@inproceedings{snavely2011discrete,
  author={D. Crandall and A. Owens and N. Snavely and D. Huttenlocher},
  title={Discrete-continuous optimization for large-scale structure from motion},
   booktitle ="IEEE International Conference on Computer Vision and Pattern Recognition",
   year = "2011",
}
 
 
 
 @inproceedings{rome09,
   author={Sameer Agarwal and Noah Snavely and Ian Simon and Steven Seitz and Richard Szeliski},
   title={Building rome in a day},
    booktitle = "International Conference on Computer Vision",
    year = "2009",
}


--new intro


@inproceedings{ArcGIS,
   author = "",
   title = "ArcGIS - Mapping and Spatial Analysis for Understanding Our World, \emph{https://www.arcgis.com/}",
   booktitle = "",
   year = "",
}

@inproceedings{USGS,
   author = "",
   title = "U.S. Geological Survey, \emph{http://www.usgs.gov/}",
   booktitle = "",
   year = "",
}

@article{GeoReg_book,
  title={Feature-based georegistration of aerial images},
  author={Sheikh, Yaser and Khan, Sohaib and Shah, Mubarak},
  journal={GeoSensor Networks},
  volume={4},
  year={2004},
  publisher={CRC Press}
}


@article{GeoReg_book,
  title={Feature-based georegistration of aerial images},
  author={Sheikh, Yaser and Khan, Sohaib and Shah, Mubarak},
  journal={GeoSensor Networks},
  volume={4},
  year={2004},
  publisher={CRC Press}
}



@article{theWayTheyMove,
	title={The way they move: Tracking multiple targets with similar appearance.},
	author={Dicle, Caglayan, Octavia Camps, and Mario Sznaier.},
	journal={Proceedings of the IEEE International Conference on Computer Vision},
	volume={},
	year={2013},
	publisher={}
}



@article{spectralMatching,
	title={A probabilistic approach to spectral graph matching.},
	author={Egozi, Amir, Yosi Keller, and Hugo Guterman.},
	journal={Pattern Analysis and Machine Intelligence, IEEE Transactions on},
	volume={},
	year={2013},
	publisher={}
}


@article{egoDailyAction,
	title={Learning to recognize daily actions using gaze. },
	author={Fathi A, Li Y, Rehg JM.},
	journal={Computer Vision–ECCV },
	volume={},
	year={2012},
	publisher={}
}


@article{egoActionFathi,
	title={Understanding egocentric activities. },
	author={Fathi A, Farhadi A, Rehg JM.},
	journal={Computer Vision (ICCV), 2011 IEEE International Conference on. IEEE},
	volume={},
	year={2011},
	publisher={}
}



@article{egoObjectDetection,
	title={Learning to recognize objects in egocentric activities. },
	author={Fathi, Alireza, Xiaofeng Ren, and James M. Rehg. },
	journal={Computer Vision and Pattern Recognition (CVPR), 2011 IEEE Conference On},
	volume={},
	year={2011},
	publisher={}
}


@article{egoVideoSummarization,
	title={Story-driven summarization for egocentric video.},
	author={Lu, Zheng, and Kristen Grauman.},
	journal={Computer Vision and Pattern Recognition (CVPR), IEEE Conference On},
	volume={},
	year={2013},
	publisher={}
}


@article{egoMobileFixedObjectDetection,
	title={Object detection and matching with mobile cameras collaborating with fixed cameras.},
	author={Alahi, Alexandre, Michel Bierlaire, and Murat Kunt.},
	journal={Workshop on Multi-camera and Multi-modal Sensor Fusion Algorithms and Applications-M2SFA2},
	volume={},
	year={2008},
	publisher={}
}



@article{egoMobileFixedMasterSlave,
	title={A master-slave approach for object detection and matching with fixed and mobile cameras.},
	author={Alahi A, Marimon D, Bierlaire M, Kunt M.},
	journal={InImage Processing, 2008. ICIP 2008. 15th IEEE International Conference},
	volume={},
	year={2008},
	publisher={}
}





@article{egoSocialInteractions,
	title={Social interactions: A first-person perspective.},
	author={Fathi, Alireza, Jessica K. Hodgins, and James M. Rehg.},
	journal={Computer Vision and Pattern Recognition (CVPR), IEEE Conference on.},
	volume={},
	year={2012},
	publisher={}
}



@article{egoFOVLocalization,
	title={Egocentric field-of-view localization using first-person point-of-view devices.},
	author={Bettadapura, Vinay, Irfan Essa, and Caroline Pantofaru.},
	journal={Applications of Computer Vision (WACV), IEEE Winter Conference on.},
	volume={},
	year={2015},
	publisher={}
}



@article{egoPredictingGaze,
	title={Predicting primary gaze behavior using social saliency fields.},
	author={Park, Hyun, Eakta Jain, and Yaser Sheikh. },
	journal={Proceedings of the IEEE International Conference on Computer Vision.},
	volume={},
	year={2013},
	publisher={}
}



@article{egoWisdomOfTheCrowd,
	title={Wisdom of the crowd in egocentric video curation.},
	author={Hoshen, Yedid, Gil Ben-Artzi, and Shmuel Peleg. },
	journal={Proceedings of the IEEE Conference on Computer Vision and Pattern Recognition Workshops. 2014},
	volume={},
	year={2014},
	publisher={}
}


@article{egoYouDoILearn,
	title={You-Do, I-Learn: Discovering Task Relevant Objects and their Modes of Interaction from Multi-User Egocentric Video.},
	author={Damen D, Leelasawassuk T, Haines O, Calway A, Mayol-Cuevas WW.},
	journal={BMVC},
	volume={},
	year={2014},
	publisher={}
}


@article{reidReimannian,
	title={Multiple-shot human re-identification by mean riemannian covariance grid. },
	author={Bak S, Corvee E, Bremond F, Thonnat M.},
	journal={InAdvanced Video and Signal-Based Surveillance (AVSS), 8th IEEE International Conference on },
	volume={},
	year={2011},
	publisher={}
}


@article{reidSDALF,
	title={Symmetry-driven accumulation of local features for human characterization and re-identification.},
	author={Bazzani L, Cristani M, Murino V.},
	journal={omputer Vision and Image Understanding.},
	volume={},
	year={2013},
	publisher={}
}

@article{reidCPS,
	title={Custom Pictorial Structures for Re-identification.},
	author={Cheng DS, Cristani M, Stoppa M, Bazzani L, Murino V. },
	journal={BMVC},
	volume={},
	year={2011},
	publisher={}
}




@article{egoHeadMotion,
	title={Head motion signatures from egocentric videos.},
	author={Cheng DS, Cristani M, Stoppa M, Bazzani L, Murino V. },
	journal={InComputer Vision--ACCV. Springer International Publishing.},
	volume={},
	year={2014},
	publisher={}
}



@article{egoYouDoILearn,
	title={Head motion signatures from egocentric videos.},
	author={Poleg Y, Arora C, Peleg S.},
	journal={BMVC},
	volume={},
	year={2014},
	publisher={}
}




@article{egoSurfing,
	title={Ego-surfing first person videos.},
	author={Yonetani, Ryo, Kris M. Kitani, and Yoichi Sato.},
	journal={Computer Vision and Pattern Recognition (CVPR), 2015 IEEE Conference on. IEEE,},
	volume={},
	year={2015},
	publisher={}
}





@article{egoWhereAmI,
	title={Where am I? Investigating map matching during self‐localization with mobile eye tracking in an urban environment.},
	author={Kiefer, Peter, Ioannis Giannopoulos, and Martin Raubal. },
	journal={Transactions in GIS 18.5},
	volume={},
	year={2014},
	publisher={}
}





@article{egoMultiTaskClustering,
	title={Egocentric daily activity recognition via multitask clustering.},
	author={Yan, Yan, et al.},
	journal={Image Processing, IEEE Transactions on},
	volume={},
	year={2015},
	publisher={}
}



@article{egoKanade,
	title={First-person vision.},
	author={Kanade, Takeo, and Martial Hebert.},
	journal={Proceedings of the IEEE 100.8 },
	volume={},
	year={2012},
	publisher={}
}



@article{egoExo,
	title={Egocentric and exocentric teleoperation interface using real-time, 3D video projection.},
	author={Ferland F, Pomerleau F, Le Dinh CT, Michaud F.},
	journal={InHuman-Robot Interaction (HRI), 2009 4th ACM/IEEE International Conference on},
	volume={},
	year={2009},
	publisher={}
}


@article{egoEvolutionSurvey,
	title={The evolution of first person vision methods: A survey. },
	author={Betancourt A, Morerio P, Regazzoni CS, Rauterberg M.},
	journal={Circuits and Systems for Video Technology, IEEE Transactions on},
	volume={},
	year={2015},
	publisher={}
}


@inproceedings{egoli2013learning,
	title={Learning to predict gaze in egocentric video},
	author={Li, Yin and Fathi, Alireza and Rehg, James},
	booktitle={Proceedings of the IEEE International Conference on Computer Vision},
	pages={3216--3223},
	year={2013}
}


@article{egoPolatsekNovelty,
	title={Novelty-based Spatiotemporal Saliency Detection for Prediction of Gaze in Egocentric Video},
	author={Polatsek, Patrik and Benesova, Wanda and Paletta, Lucas and Perko, Roland},
	publisher={IEEE}
}


@article{Borji2014look,
	title={What/where to look next? Modeling top-down visual attention in complex interactive environments},
	author={Borji, Ali and Sihite, Dicky N and Itti, Laurent},
	journal={Systems, Man, and Cybernetics: Systems, IEEE Transactions on},
	volume={44},
	number={5},
	pages={523--538},
	year={2014},
	publisher={IEEE}
}

\end{document}